\documentclass{article}

\usepackage{adjustbox}

\usepackage{graphicx}
\usepackage{booktabs}
\usepackage{multirow}
\usepackage{placeins}
\usepackage[pagebackref,breaklinks,colorlinks]{hyperref}
\usepackage{arxiv}
\usepackage{amsmath}

\usepackage[utf8]{inputenc} 
\usepackage[T1]{fontenc}    
\usepackage{hyperref}       
\usepackage{url}            
\usepackage{booktabs}       
\usepackage{amsfonts}       
\usepackage{nicefrac}       
\usepackage{microtype}      
\usepackage{lipsum}		
\usepackage{graphicx}
\usepackage{doi}
\newcommand\chapter{\clearpage
                    \thispagestyle{empty}%
                    \global\@topnum\z@
                    \@afterindentfalse
                    \secdef\@chapter\@schapter}
\newcommand{\extendeddatafigs}{%
\renewcommand{\figurename}{Supplementary Fig.}
  \setcounter{figure}{0}
  \let\oldthefigure\thefigure
  
  \renewcommand{\thefigure}{\oldthefigure}
  \let\oldchapter\chapter
  \renewcommand{\chapter}{
    \let\thefigure\oldthefigure
    \let\chapter\oldchapter
    \oldchapter
}
}
\newcommand{\extendeddatatab}{%
\renewcommand{\tablename}{Supplementary Table}
  \setcounter{table}{0}
  \let\oldthetable\thetable
  \renewcommand{\thetable}{\oldthetable}
  \let\oldchapter\chapter
  \renewcommand{\chapter}{
    \let\thetable\oldthetable
    \let\chapter\oldchapter
    \oldchapter
}
}

\newcommand{\tran}{^\top}

\newcommand{\real}{\mathbb{R}}

\newcommand{\mif}{\textrm{if }}
\newcommand{\other}{\textrm{otherwise}}

\title{Get Your Embedding Space in Order: Domain-Adaptive Regression for Forest Monitoring}

\author{ \href{https://orcid.org/0000-0001-9583-7230}{\includegraphics[scale=0.06]{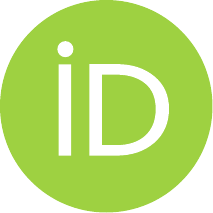}\hspace{1mm}Sizhuo Li}\thanks{Equal contributions} \\
	University of Copenhagen\\
	Øster Voldgade 10, 1350 København K, Denmark \\
 Université Paris-Saclay\\
 Gif-sur-Yvette 91190, France\\
	\texttt{sizli@ign.ku.dk} \\
	\And
	\href{https://orcid.org/0000-0002-8135-1341}{\includegraphics[scale=0.06]{orcid.pdf}\hspace{1mm}Gominski Dimitri}\footnotemark[1] \\
	University of Copenhagen\\
	Øster Voldgade 10, 1350 København K, Denmark \\
	\texttt{dg@ign.ku.dk} \\
 \And
	\href{https://orcid.org/0000-0001-9531-1239}{\includegraphics[scale=0.06]{orcid.pdf}\hspace{1mm}Martin Brandt}\\
	University of Copenhagen\\
	Øster Voldgade 10, 1350 København K, Denmark \\
	\texttt{mabr@ign.ku.dk} \\
 \And
	\href{https://orcid.org/0000-0002-9709-0633}{\includegraphics[scale=0.06]{orcid.pdf}\hspace{1mm}Xiaoye Tong}\\
	University of Copenhagen\\
	Øster Voldgade 10, 1350 København K, Denmark \\
	\texttt{xito@ign.ku.dk} \\
 \And
	\href{https://orcid.org/0000-0001-8560-4943}{\includegraphics[scale=0.06]{orcid.pdf}\hspace{1mm}Philippe Ciais}\\
	CEA, CNRS, UVSQ, Université Paris-Saclay\\
 Gif-sur-Yvette 91190, France\\
	\texttt{philippe.ciais@cea.fr} \\
}

\renewcommand{\headeright}{}
\renewcommand{\undertitle}{}

\hypersetup{
pdftitle={Get Your Embedding Space in Order: Domain-Adaptive Regression for Forest Monitoring},
pdfsubject={q-bio.NC, q-bio.QM},
pdfauthor={Sizhuo Li, Dimitri Gominski, Martin Brandt, Xiaoye Tong, Philippe Ciais},
pdfkeywords={Dataset, Regression, Remote sensing, Domain adaptation},
}

\begin{document}
\maketitle

\begin{abstract}
	Image-level regression is an important task in Earth observation, where visual domain and label shifts are a core challenge hampering generalization. However, cross-domain regression within remote sensing data remains understudied due to the absence of suited datasets. We introduce a new dataset with aerial and satellite imagery in five countries with three forest-related regression tasks\footnote[2]{Dataset and code available here: \href{http://dgominski.github.io/drift/}{dgominski.github.io/drift/}}. To match real-world applicative interests, we compare methods through a restrictive setup where no prior on the target domain is available during training, and models are adapted with limited information during testing. Building on the assumption that ordered relationships generalize better, we propose manifold diffusion for regression as a strong baseline for transduction in low-data regimes. Our comparison highlights the comparative advantages of inductive and transductive methods in cross-domain regression.

\end{abstract}


\section{Introduction}
\label{sec:intro}

Distribution shifts between the training data and testing data are ubiquitous in computer vision. While it is generally accepted that a model trained on sufficient data has some ability to generalize to new target data, in practice visual variations induce significant performance drops \cite{Beery_2018_ECCV}. 

A good example can be found in vegetation mapping and monitoring tasks. Biodiversity monitoring, carbon tracking, resource management, and food security analysis heavily depend on a regular assessment of forest resources through important parameters such as tree counts or height. A practical solution is to train image models to regress such values from satellite or aerial imagery \cite{chen_transformer_2022, li2023deepbiomass}. However, models trained on remote sensing imagery are usually limited to one area and perform poorly on others, due to different spectral ranges, ground resolutions, or visual patterns \cite{wang2024trustworthy, Marsocci_2023_CVPR, Voulgaris_2023_CVPR}.

Data availability and downstream applications define the scope of models (regional \cite{li_rse_2020} -- national \cite{mugabowindekwe_nation-wide_2023} -- continental \cite{brandt_unexpectedly_2020}), but they don't necessarily match the visual and semantic domains defined by biomes and environmental conditions. Intuitively, a model trained on a dataset should perform relatively well on datasets that are spatially close and visually similar, although it remains unclear to what extent.

To better assess generalization performance and go towards universal models that can be trained in countries with high data availability and applied to any country without prior knowledge or retraining, we introduce the DRIFT dataset for universal, low-data domain-adaptive regression.

\begin{figure}[t]
    \centering
    \includegraphics[trim={2.5cm, 0cm, 0.5cm, 0cm},clip,width=0.7\textwidth,keepaspectratio]{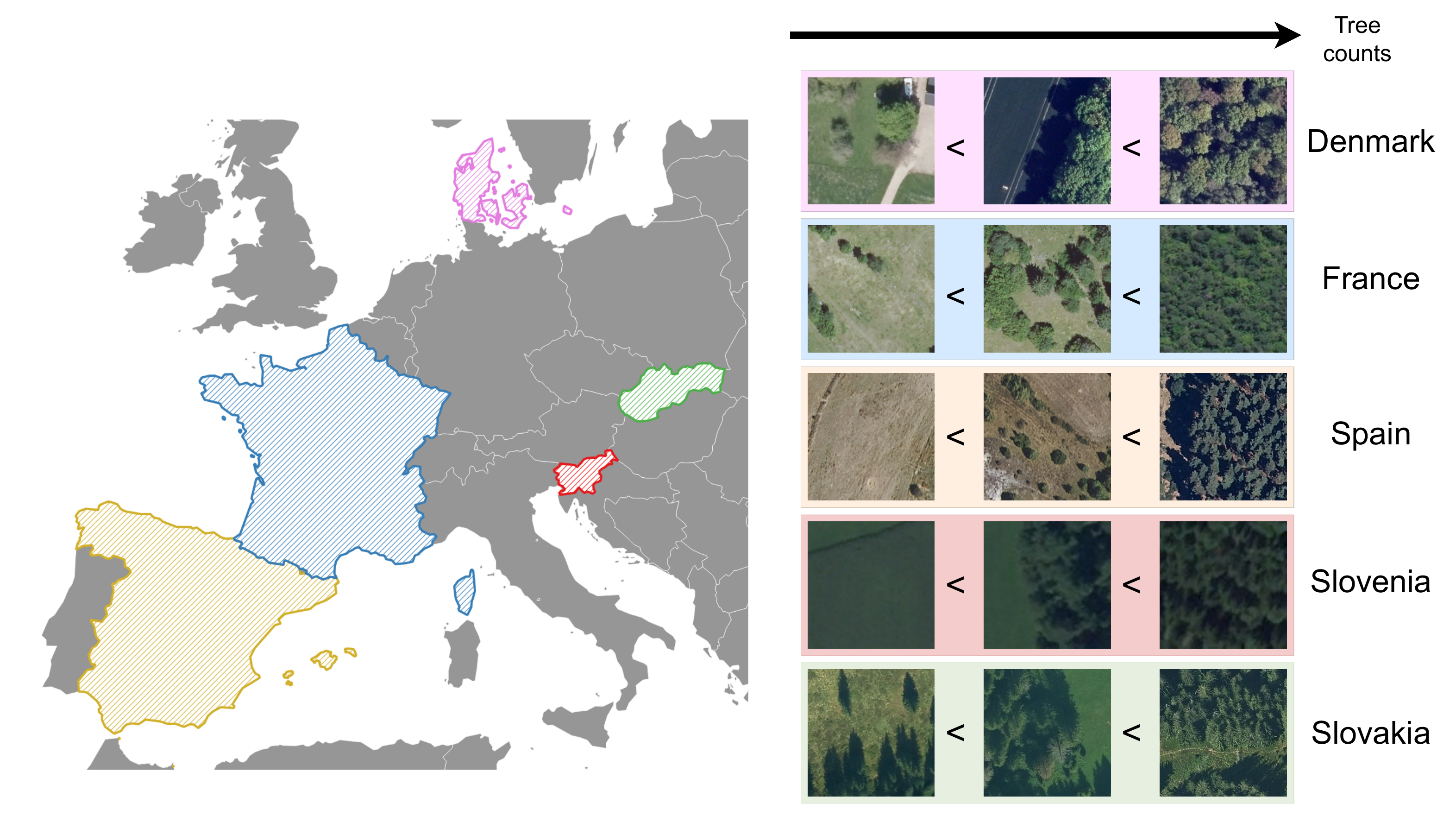}
    \caption{Vegetation regression across countries. We hypothesize that order relations generalize better through visual and label domains than direct regression. Predicting tree counts directly requires domain-specific knowledge about local species, whereas predicting ordered chains remains intuitively easy.}
    \label{fig:idea}

\end{figure}

Building on recent advances in order learning, we hypothesize that order relationships generalize well across domains (Fig.~\ref{fig:idea}). We enforce a well-ordered embedding space in the source domain, then adapt to target domains in low-data setups. Notably, this strong regularization during training does not hamper source-domain performance \textbf{and} creates a clean embedding space to propagate similarities at test-time, which opens the way for advanced low-data inference in the form of manifold diffusion. Our contributions are as follows:
\begin{itemize}
    \item We publish a large-scale dataset across five countries in Europe for domain-adaptive regression, using both aerial and satellite imagery, with a panel of tasks related to forest monitoring. 
    \item We compare inductive and transductive methods across datasets and tasks, in a low-shot source-free domain adaptation setup. We show that transductive methods have an advantage when the domain gap becomes predominant.
    \item We introduce a simple but effective baseline for domain-adaptive regression. We propose a generalization of graph diffusion, a common technique for semi-supervised classification, to image-level regression. Our experiments demonstrate the applicative potential and superiority over common baselines. 
\end{itemize}

\section{Related work}

\subsection{Order learning}

The seminal work of \cite{lim_order_2020} introduced order learning as an alternative to direct regression, with the motivation that in certain cases, it is easier to compare images (this person is older than this other person) than to associate them independently with a scalar. Subsequent works built on this idea with iterative predictions \cite{shin_moving_2022}, partial ordering \cite{lee_order_2022} and further constraints on the geometry of the embedding space \cite{lee_geometric_2022}. Our work builds on the latter. In a cross-domain setup, we hypothesize that the highly constrained embedding space with geometric constraints in the source domain generalizes better to other domains.

\subsection{Cross-Domain Regression}

Existing works have tackled domain adaptation through three main setups, depending on the available information about the target domain.

\textit{Unsupervised Domain Adaptation} trains on unlabeled images of the target domain. Some works have focused on aligning features of the source and target domain with dedicated mechanisms to limit scale \cite{chen_representation_2021} or embedding structure \cite{nejjar_dare-gram_2023} shifts, so that the regressor maintains its performance on the target domain. Adversarial training \cite{mathelin_adversarial_2021} has been proven effective as a tool to learn a domain shift metric.

\textit{Semi-Supervised Domain Adaptation}, in contrast, assumes that some labeled images are available in the target domain. A common approach is to weight training points to minimize the performance gap \cite{wang_transfer_2019}, or as a form of boosting \cite{pardoe_boosting_2010}. Few-shot domain adaptation can be considered a special case of semi-supervised adaptation, where the number of target domain labeled samples is very limited. A good example is the work of Teshima et al. \cite{teshima_few-shot_2020}, where authors generate synthetic target domain samples with the assumption that generative modeling offers more invariance. 

These setups have limited applicability in many real-world problems due to the fact that utilizing labeled or unlabeled data from the target domain during training requires specific training for each target domain.

\textit{Source-Free Domain Adaptation} removes the liberty of modifying the source domain training phase, which is not always possible in real settings. Models are only adapted at test-time, when applied on the target domain. In the context of image classification, the popular SHOT \cite{liang_we_2020} method introduced information maximization and self-supervised learning as promising ideas for label-free target domain fine-tuning. If a few labeled samples are available, simple fine-tuning on the target domain with suited regularization remains a strong baseline \cite{lee_few-shot_2023}. We are not aware of works exploring this for regression.

Here, we aim at building a model that can be quickly adapted to any target domain, without any prior knowledge of that domain. As such, we prepare the model for generalization on the source domain, only with source domain data. At test-time, we explore low-data configurations, with very limited information available on the target domain for quick adaptation. This setup has been labeled as \textbf{universal source-free domain adaptation} \cite{kundu_universal_2020} in the context of classification.

\subsection{Image-Level Regression in Earth Observation}

The increasing availability of vast and diverse collections of remote sensing imagery and datasets presents both opportunities and challenges for computer vision tasks. To date, most works utilizing remote sensing imagery focus on image classification and semantic segmentation \cite{akiva_self-supervised_2022, robinson_large_2019, beery_auto_2022}. For example, to count trees, some works rely on the detection of individual trees \cite{li_deep_2023, brandt_unexpectedly_2020}, but manual annotation relies on expert knowledge and is extremely time-consuming. Given the variations in ground resolution, image quality, and geographic areas, small objects and occlusion phenomena can make exhaustive annotation virtually impossible. Other labels on woody biomass or canopy height cannot be manually labeled. Therefore, labels often have to be sourced from field measures \cite{li_forest_2020, deng_comparison_2016}, or derived from low-resolution dedicated sensors such as GEDI \cite{lang_high-resolution_2023, fayad_vision_2023}, and remain noisy due to the varying viewing angles, technical inaccuracies in positioning systems \cite{girard_noisy_2019} or temporal mismatch \cite{sumbul_label_2023, burgert_effects_2022}. As such, accurate labels are often found at a lower resolution than the ground resolution of the image. This makes image-level regression a core task in computer vision for Earth observation, as it offers a way to directly predict the information at the available resolution \cite{drivendata_biomassters_nodate, chen_transformer_2022}. To our knowledge, there is no existing dataset for cross-domain image-level regression using sub-meter resolution remote sensing imagery (Supplementary Tables 2 and 3).

\section{DRIFT: Domain-Adaptive Regression for Forest Monitoring across Countries}

We introduce the DRIFT dataset (Domain-adaptive Regression for Image-level Forest moniToring), including 25k images in five European countries from aerial and nanosatellite imagery, with three target variables to predict for each image:
\begin{enumerate}
    \item Canopy height: average height value for pixels containing woody vegetation.
    \item Tree count: number of overstory (visible from an overhead perspective) trees in the images. 
    \item Tree cover fraction: percentage of the image being covered by overstory tree crowns.  
\end{enumerate}

\textit{Dataset Collection.} We built the dataset from national aerial orthophotography with various ground sampling distances (GSD) for Denmark (DK), Spain (SP), France (FR), and Slovakia (SK). All aerial images contain three spectral channels: Red (R), Green (G), and Blue (B). For Slovenia (SI), we used very high-resolution satellite imagery from commercial nanosatellites (SkySat). 
We acquired the orthorectified surface reflectance products at a spatial resolution of 50cm. 
Characteristics of each subset in the DRFIT are summarized in Table~\ref{tab:drift_characteristics}.
Refer to Supplementary section 1 for extended details regarding the dataset curation, data origins, and visual examples. 

\begin{table}[]
    \centering
    \caption{DRIFT Dataset characteristics. GSD and Height in meters, Cover in percents.}
    \label{tab:drift_characteristics}
    \begin{tabular*}{\textwidth}{@{}l@{\extracolsep{\fill}}*{8}{c}@{}}\toprule
        \
        \multirow{2}{*}{Country} & \multirow{2}{*}{\# Images} & \multirow{2}{*}{Patch size} & \multirow{2}{*}{Sensor} & \multirow{2}{*}{GSD} & \multirow{2}{*}{Timeframe} & \multicolumn{3}{c}{Range} \\
         & & & & & & Height & Count & Cover \\
        \midrule
Denmark & 13094 & 200 & aerial & 0.2 & 2018 & $[0, 38]$ & $[0, 192]$ & $[0, 100]$\\
        France & 10298 & 250 & aerial & 0.2 & 2018-20 & $[0, 28]$ & $[0, 204]$ & $[0, 100]$\\
        Slovakia & 3129 & 250 & aerial & 0.2 & 2021 & $[1, 38]$ & $[0, 148]$ & $[0, 100]$\\
        Spain & 1150 & 200 & aerial & 0.25 & 2020 & $[0, 29]$ & $[0, 135]$ & $[0, 100]$\\
        Slovenia & 1903 & 100 & satellite & 0.5 & 2021 & $[0, 36]$ & $[0, 91]$ & $[0, 100]$\\
        \bottomrule
    \end{tabular*}
\end{table}

\textit{Dataset Annotation.} Aerial LiDAR is an accurate source of 3D structure information to characterize trees across various landscapes \cite{weinstein_benchmark_2021, knapp_linking_2018, kalinicheva_multi-layer_2022}. With the high-resolution height information, trees can be separated from other objects and measured in height and crown width. We first rasterized the LiDAR data to Canopy Height Models (CHMs), where each pixel indicates the top height of vegetation at this location, if there is any. Image-level canopy height was calculated by averaging all non-zero height values in the image, and tree cover fraction was extracted with 5-meter height thresholding to aggregate areas covered by vegetation taller than 5m \cite{fao_tree_def}. Regarding tree counts, we used a public imagery-based tree counting model for Denmark, which was trained using point supervision with 22k tree annotations \cite{li2023deep}. For the rest, we processed CHMs to extract tree crown centers with local peak identification over a sliding window with adaptive size \cite{roussel_lidr_2020}. CHMs were preprocessed with a conditional filter for higher separability of low foliage or artifacts \cite{deng_comparison_2016}. We adjusted image-level counts with linear correction. All hyperparameters were fit on a small labeled set ($\approx1$k trees) for each country, and we report the 3-fold $R^2$ score on random train/validation splits for correction in Supplementary Table 1.

\textit{Challenges.} The DRIFT dataset includes significant shifts between label and visual distributions (see Fig. \ref{fig:challenge}) due to sensor and area differences (see distribution of biomes in Supplementary Fig. 2). Moreover, within a tree species, individuals can exhibit different behaviors depending on their direct environment and living climate, which introduces concept drift: visually similar trees can vary in height \cite{jucker_allometric_2017}. Such variations can also be influenced by sensor properties such as the viewing angle.

\begin{figure}[t]
    \centering
    \includegraphics[width=\textwidth,keepaspectratio]{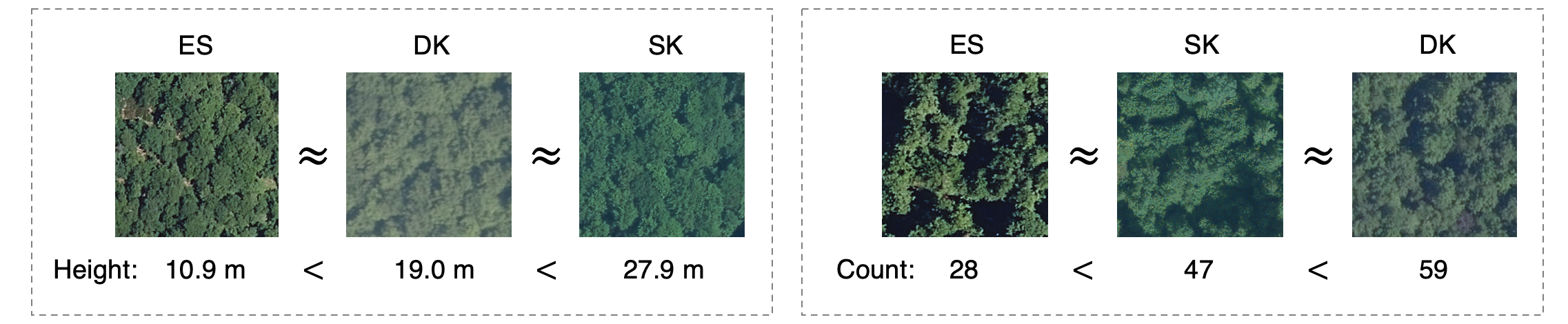}
    \caption{Challenging examples in the DRIFT dataset: despite similar visual content, images can have different values in the label space. More visual examples can be found in Supplementary Fig. 1.}
    \label{fig:challenge}
\end{figure}

\begin{figure}[t]
    \centering
    \includegraphics[trim={1.2cm, 0cm, 0cm, 0cm},clip,width=0.65\textwidth,keepaspectratio]{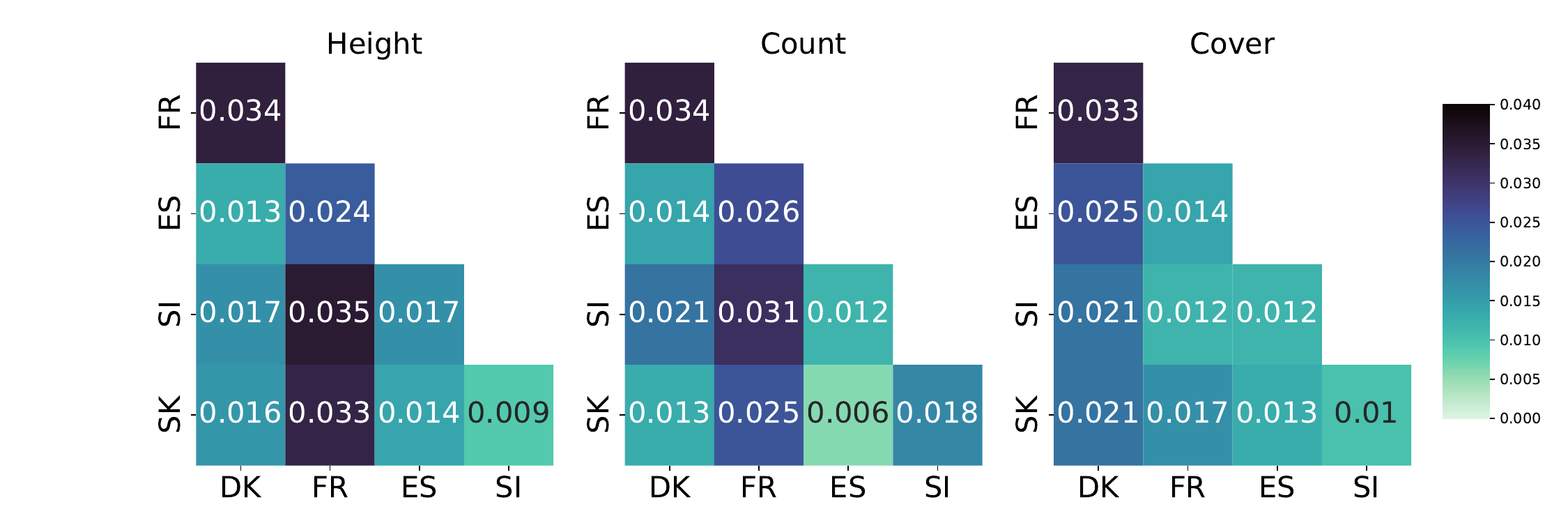}
    \caption{Label shift varies across countries and tasks. We plot Wasserstein distances between label distributions in country subsets to indicate the level of label shift.}
    \label{fig:labelshift}
\end{figure}

\textit{Label Shift.} The DRIFT dataset exhibits a substantial shift in label distribution across countries. We plot the label shift in Fig. \ref{fig:labelshift} between all possible pairs of annotation subsets in DRIFT. 

\textit{Regression Tasks.} Canopy height, tree count, and tree cover heavily depend on tree species, local climate, and landscape characteristics. Height prediction from optical images is inherently challenging due to limited visual clues about depth from an aerial viewpoint, especially if the ground is not visible, but it has been shown possible using satellite imagery \cite{lang_high-resolution_2023, liu_overlooked_2023}. Counts depend on the model's interpretation of individual crowns, which is subject to extreme variations even within a species \cite{jucker_allometric_2017}. Cover fraction is intuitively the easiest, as the model only needs to separate trees from background pixels. The difficulty depends on the level of similarity between short vegetation ($<5$m) and trees.

\textit{Proposed Splits.} We propose to use two countries (Denmark and France) with abundant high-quality CHM data as source domains for training. For testing, we apply models on three countries (Slovakia, Slovenia, Spain) representing different target domains. The Slovenia subset allows a comparison with another type of sensor (nanosatellites), and the Spain subset introduces a much drier ecosystem compared to the others. The France and Denmark subsets have different label distributions (Fig. \ref{fig:labelshift}), which introduces variation in the comparison of models trained on them. 

\textit{Evaluation.} We propose an evaluation framework for DRIFT that follows common constraints and represents real-world application contexts. First, we consider that no knowledge about the target domain is available during training. A model that can generalize to multiple areas is more valuable than models that are specific to a source/target dataset combination. As such, there is a high interest in going towards "universal" models that do not build on assumptions about their target domain. In a real-world scenario, a model that reached high performance on its source domain will be archived and considered for generalization, preferably with no or little supporting information about the target domain, and no retraining. Second, we consider that some knowledge about the target domain can be accessed at test-time. Following the terminology introduced in \cite{kundu_universal_2020}, we propose to follow a setup of universal source-free domain adaptation for regression, in two stages. In the Procurement stage, or source domain training, models are trained and prepared for generalization. 
During the Deployment stage, or target domain testing, models are adapted using a few labeled examples. In the following, we refer to this setup as $N$-shot adaptation.

\section{Method}

\begin{figure}[t]
  \centering
  \includegraphics[trim={0.8cm, 0cm, 0.8cm, 0cm},clip,width=0.85\textwidth]{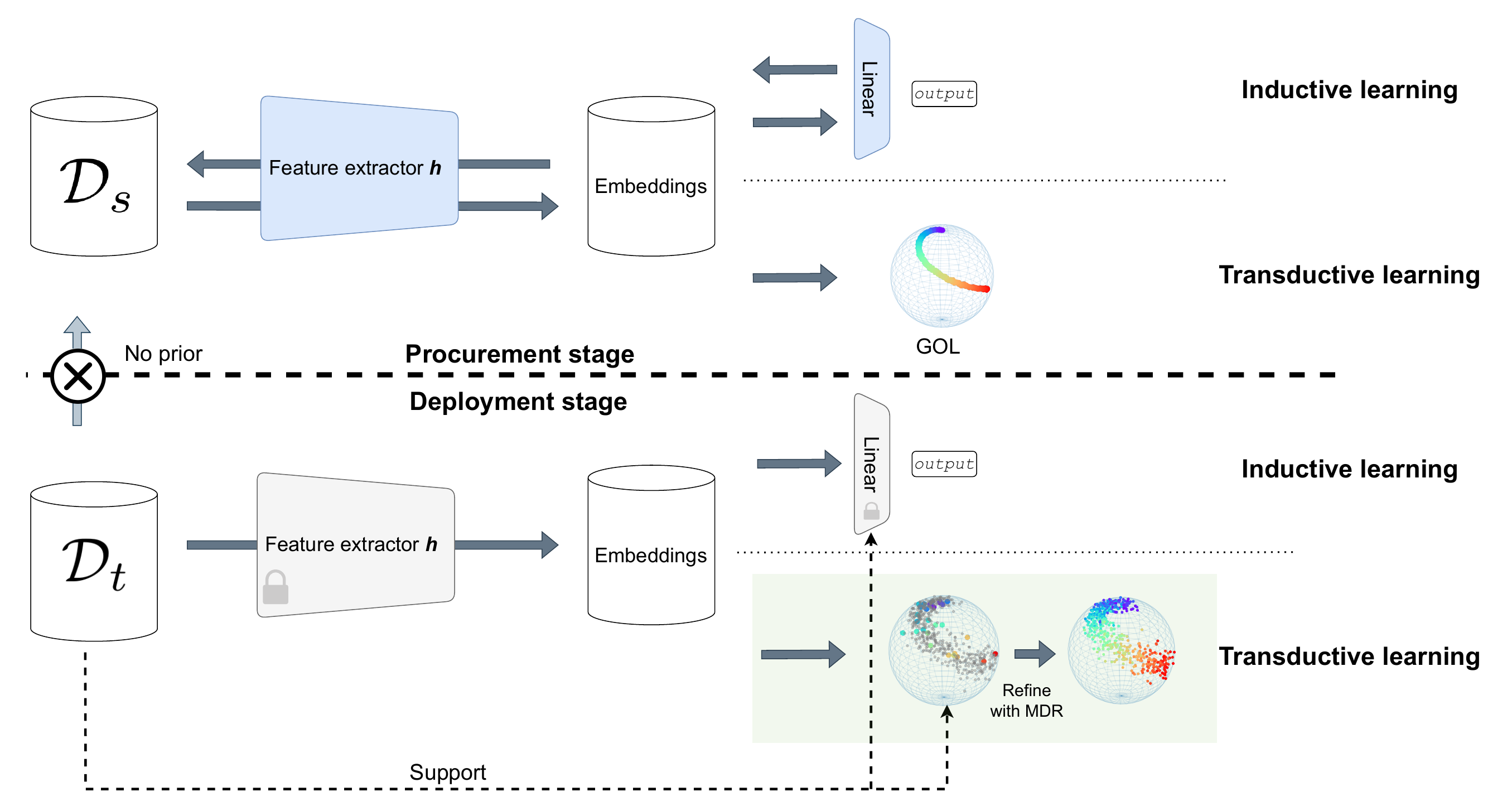}
  \caption{Overview of our universal, source-free domain adaptation framework. The feature extractor (and linear layer for inductive regression) is trained on the source domain $\mathcal{D}_s$ during the Procurement stage, then applied on the target domain $\mathcal{D}_t$ during the Deployment stage. We do not use any prior regarding the target domain when training on the source domain, and do not assume sufficient data is available for training on the target domain. On the target domain, a few labeled examples are available to calibrate or fine-tune inductive methods, and support our proposed manifold diffusion for regression (MDR), a transductive approach.} 
  
  \label{fig:overview}
\end{figure}

We explore the potential of ordered embeddings and simple transductive learning on the cross-domain challenges of the DRIFT dataset (see Fig. \ref{fig:overview}). We train the embedding extractor on the source domain, using geometric order learning to enforce structure. We extract embeddings on the target domain, and perform adaptation for regression using a few labels. Optionally, we refine predictions to take into account local and global manifold structure with diffusion.

\subsection{Preliminaries}

We first introduce notations for the problem of few-shot domain adaptation for regression.

Let $\mathcal{D}_s=\{(\textbf{x}_i^s,y_i^s)\}$ and $\mathcal{D}_t=\{(\textbf{x}_i^t,y_i^t)\}$ denote the source and target domains  respectively. $y^s$ and $y^t$ take continuous values in $\real$, and we note that their distributions are different. Notably, it is common in vegetation mapping that their support $[y^s_{min}, y^s_{max}]$ and $[y^t_{min}, y^t_{max}]$ differs. For example, $y^s_{max} \neq y^t_{max}$ in height mapping due to different species.

At test-time, we assume that some information about the target domain is accessible, in the form of a few labeled samples, which we refer to as the target domain support set $\mathcal{D}_{ts} \subset \mathcal{D}_t$. Similarly to the common practice in classification \cite{snell_prototypical_2017, finn_model-agnostic_2017}, we divide the target label set into $k_r$ equally spaced value groups, and construct the target domain support set $D_{ts}$ by randomly drawing $N$ samples from each group, forming a $k_r$-way $N$-shot adaptation setup.

A simple baseline method would be to train a regressor $f(\textbf{x}) = y$ on $\mathcal{D}_s$ and apply it to $\mathcal{D}_t$. Due to domain shifts, this performs generally poorly. Instead, we consider a feature extractor $h(\textbf{x}) = \mathbf{v}$, with $\mathbf{v} \in \real ^d$ and $\lVert \mathbf{v} \lVert_2 = 1$, and frame the problem as cross-domain transduction: $y^t$ is inferred from the relationships between the embeddings in $\mathcal{D}_t$.

Geometric order learning (GOL) \cite{lee_geometric_2022} enforces the notion of order in the embedding space, such that the direction and distance between embeddings reflect the relations between their labels. The framework is based on $M$ reference points that are shared between embeddings of similar values. Consider $\mathbf{v}_a$ and $\mathbf{v}_b$ and the corresponding reference points $r_{\theta(a)}$, $r_{\theta(b)}$, with $\theta$ the function assigning a sample to its reference point. GOL trains with the following composite loss function:
\begin{equation}
    L = L_o + L_m + L_c
\end{equation}
where 
\begin{itemize}
    \item $L_o$, the order loss, aligns the embeddings with the direction of their label relation. For the $r_{\theta(a)} < r_{\theta(b)}$ case, $L_o = -\log e^{\overrightarrow{u_+} \cdot \overrightarrow{u_{ab}}} / (e^{\overrightarrow{u_+} \cdot \overrightarrow{u_{ab}}} + e^{\overrightarrow{u_-} \cdot \overrightarrow{u_{ab}}})$ with $\overrightarrow{u_+} = (r_{\theta(b)} - r_{\theta(a)}) / \lVert r_{\theta(b)} - r_{\theta(a)} \lVert$ the forward direction vector between the reference points of $a$ and $b$, $\overrightarrow{u_-} = (r_{\theta(a) - 1} - r_{\theta(a)}) / \lVert r_{\theta(a) - 1} - r_{\theta(a)} \lVert$ the backward direction vector, and $\overrightarrow{u_{ab}} = (\mathbf{v}_b - \mathbf{v}_a) / \lVert \mathbf{v}_b - \mathbf{v}_a \lVert$. When $r_{\theta(a)} > r_{\theta(b)}$, the corresponding formula is obtained symmetrically.
    \item $L_m$, the metric loss, makes embedding space distances reflect label differences. When $r_a = r_b$, $L_m = \sum_{i=0}^{M-1} \max (| d(r_i, \mathbf{v}_a) - d(r_i, \mathbf{v}_b) | - \gamma, 0)$, with $d$ the Euclidean distance. When $r_{\theta(a)} < r_{\theta(b)}$, $L_m = \sum_{i=0}^{a} \max (| d(r_i, \mathbf{v}_a) - d(r_i, \mathbf{v}_b) | + \gamma, 0) + \sum_{j={b}}^{M-1} \max (| d(r_j, \mathbf{v}_b) - d(r_j, \mathbf{v}_a) | + \gamma, 0)$. That is, all reference points inferior to $r_{\theta(a)}$ pull $\mathbf{v}_a$ closer, while all reference points superior to $r_{\theta(b)}$ push $\mathbf{v}_a$ away, and the opposite for $\mathbf{v}_b$. The formula for $r_{\theta(a)} > r_{\theta(b)}$ is obtained symmetrically.
    \item $L_c$, the center loss, places reference points at the center of embeddings with similar values. $L_c = d(r_{\theta(a)}, \mathbf{v}_a) + d(r_{\theta(b)}, \mathbf{v}_b)$.
\end{itemize}
We invite interested readers to refer to the original paper for more details.

Lee et al. \cite{lee_geometric_2022} report that GOL creates a well-ordered embedding space, which allows prediction with a simple distance-based k-nearest neighbors (kNN) assignment to estimate continuous values.

\subsection{Manifold Diffusion for Regression}

We introduce Manifold Diffusion for Regression (MDR) as an optional post-processing step to refine the predictions. Manifold diffusion \cite{iscen_efficient_2017, donoser_diffusion_2013, zhou_learning_2003} is a popular transductive method in semi-supervised learning and image retrieval to exploit the structure of the embedding space for better class predictions. Here, we modify it for regression in the context of cross-domain adaptation.

We start by introducing the diffusion framework originally proposed in \cite{zhou_learning_2003}. Assuming a set $X = \{\textbf{x}_1, \textbf{x}_2, ..., \textbf{x}_n\}$ and $\mathbf{v}_i=h(\textbf{x}_i)$, the affinity matrix $W \in \real ^{n \times n}$ is constructed by calculating the pairwise cosine similarity between samples: 

\begin{equation}
	W_{ij}:=
	\begin{cases}
		[\mathbf{v}_i\tran \mathbf{v}_j]_+^\gamma, & \mif i \ne j \wedge \mathbf{v}_i \in \text{NN}_{k} (\mathbf{v}_j)\\
		0, & \other
	\end{cases}
\label{eq:affinity}
\end{equation}

with $\text{NN}_{k}$ the set of $k$ nearest neighbors in $X$, and $\gamma$ a hyperparameter. $W$ is non-negative, sparse, and has zero diagonal.

The symmetric normalized form is computed:

\begin{equation}
    W_n:= D^{-1/2}WD^{-1/2}
\label{eq:w}
\end{equation}

where $D$ is the diagonal matrix $D:=diag(W1_n)$ with the row-wise sums of $W$. The normalized graph Laplacian of $W$ is defined as $L:= I_n - W_n$. Assuming that we have label information for some of the points, we can iteratively update the graph with $F(t + 1) = \alpha W F (t) + (1 - \alpha) Y$, where $Y \in \{0, 1\}^{n \times c}$ is the one-hot encoded label matrix for $c$ classes. Zhou et al. \cite{zhou_learning_2003} noted that this is equivalent to solving the linear system: $F^{*}:=(I_n - \alpha W_n)^{-1}Y$. The final class prediction for a sample $i$ is obtained as $\textsc{argmax} (F_{i})$.

We replace the label matrix $Y$ with the support matrix $S \in \{0, 1\}^{n \times k}$, the one-hot encoded matrix for each of the $k$ labeled samples. $S_{i,j} = 1$ for the first $k$ rows and columns, and zero elsewhere. After solving for $S^* = (I_n - \alpha W_n)^{-1}S$, we obtain an updated similarity score between an image and each of the support elements. $S^*$ captures the structure of the embedding space and creates a refined similarity measure taking into account local and global context. 

We output the final regression estimate as the weighted sum of the $k_v$ highest values in $S^*$ and the known labels. Note that this step heavily relies on a well-ordered embedding space: predictions will be unstable if the embedding space does not maintain the order of the one-dimensional continuous label space. 

\section{Experiments}

\subsection{Implementation details}
During the Procurement stage, we train a ViT-B/16 \cite{dosovitskiy_image_2021} feature extractor with GOL with $M = k_r = 5$ and equal interval between label groups. For canopy height, the upper bound is set to 25m, and each group spans over 5m. For count, we use 0 as the lower bound and 90 (Denmark) or 70 (France) as the upper bound and split accordingly. For cover, we use 20\% as the lower bound and 100\% as the upper bound.
We weight the order loss $L_o$, metric loss $L_m$, and center loss $L_c$ with a factor of 1, 66, and 33, respectively, for similar convergence rates.

\begin{figure}[h!]
  \centering
  \includegraphics[height=7.9cm]{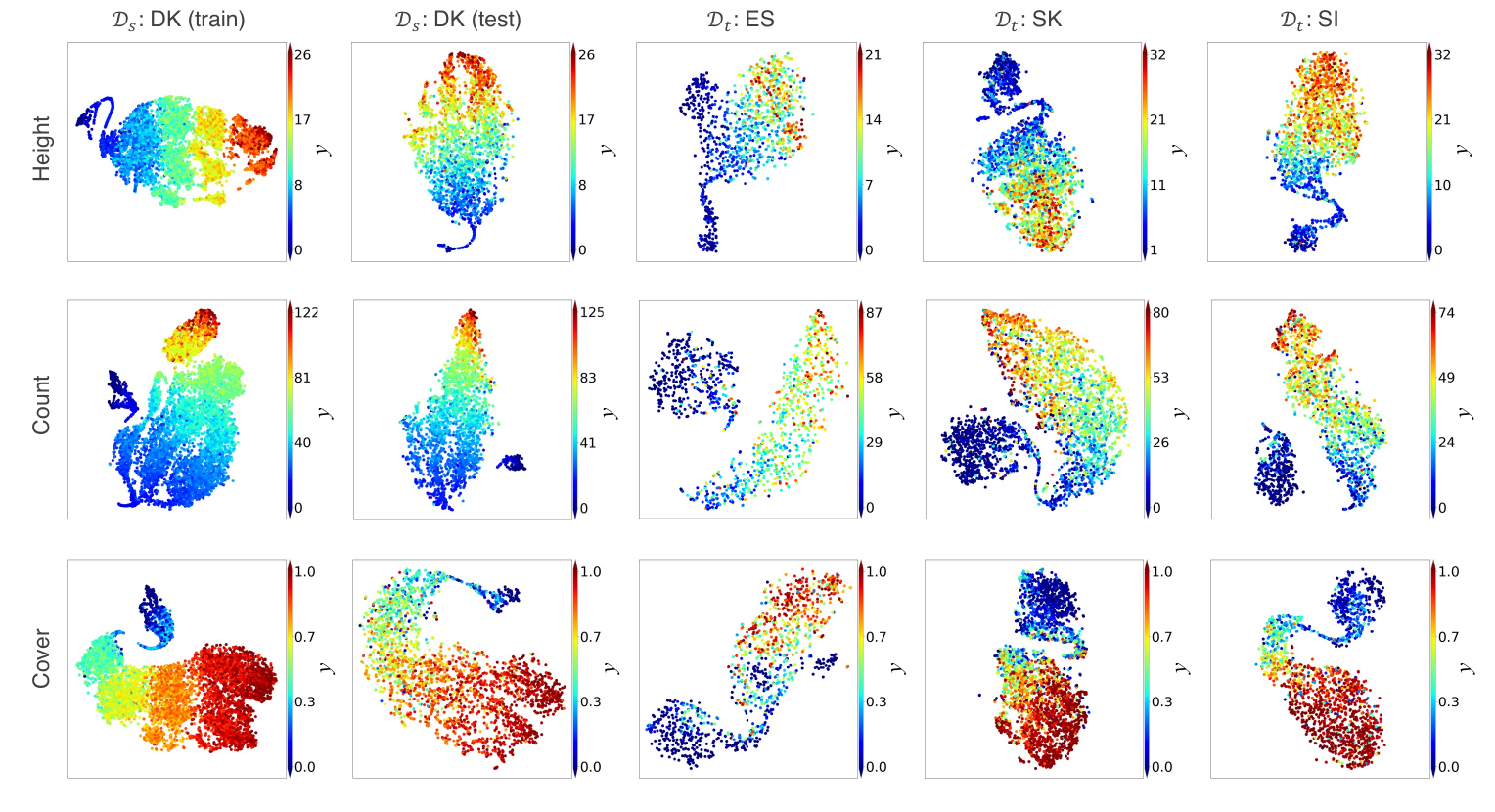}
  \caption{Embeddings retain ordered relations across domains (t-SNE visualization). From left to right: source domain training data, source domain testing data, and three target domains, respectively.}
  \label{fig:embedding}
\end{figure}

During the Deployment stage, we pick the 1\% and 99\% percentile values of each target variable as the lower and upper bound and split the entire dataset into 5 groups. 
 
We implement MDR in a few-shot setup. We use standard hyperparameters that are reported to work well across different setups ($\gamma = 3$, $\alpha = 0.99$) \cite{iscen_label_2019}, and use the entire support set for diffusion, setting $k=N$ and assigning the final values with $k_v = 2N$, with the effects of $k_v$ shown in Supplementary Fig. 6. We refer to this approach as GOL+MDR.

\subsection{Qualitative Analysis}

We plot projections of the embedding space in the source and target domains across tasks on the DRIFT dataset in Fig.~\ref{fig:embedding}. Label order in the embedding space is roughly maintained across domains, but does not necessarily remain in a single well-connected cluster with a clear direction of label space increase. Consequently, kNN label assignment can pick up noisy neighbors from adjacent clusters and wrongly predict.

\begin{figure}[]
  \centering
  \includegraphics[trim={1.6cm, 0cm, 1.3cm, 0cm},clip,width=0.85\linewidth]{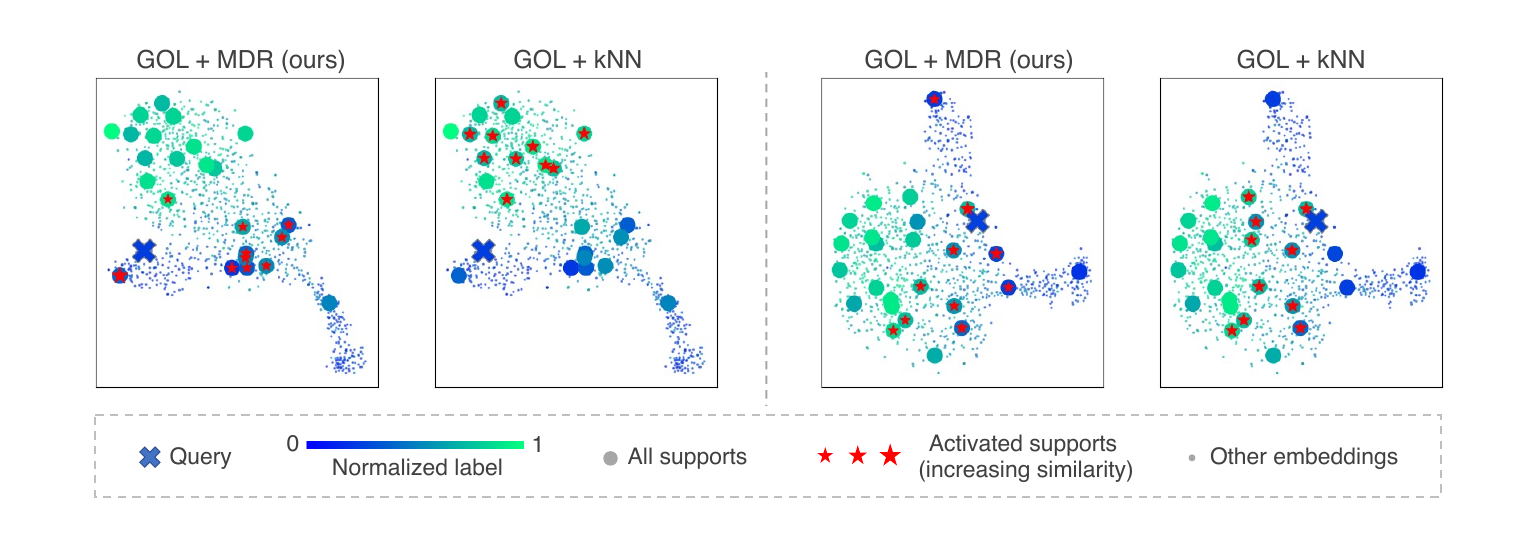}
  \caption{Top 10 support samples activated by GOL+MDR and GOL+kNN in 5-shot setup (t-SNE visualization). Closer colors indicate higher adjacency in the label space.
  }
  \label{fig:diffusion}
\end{figure}

We plot the effect of MDR on DRIFT examples in Fig.~\ref{fig:diffusion}. MDR activates more relevant support samples than kNN, i.e., the examples picked up for assigning the prediction are closer to the true value. This comes from taking into account both global and local manifold structure, rather than only the immediate neighbors. Assuming that embeddings remain overall ordered on the target domain, MDR captures higher-order similarities and discards potentially confusing support samples in the target domain. Example predictions can be found in Supplementary Fig. 3 and 4.

\subsection{Quantitative Analysis}

\subsubsection{Few-Shot Evaluation.} We compare our approach with common baselines and state-of-the-art methods in Table~\ref{tab:quantitative_fewshot}. We compare against regression on the target domain without adaptation (Regression), regression on the target domain with linear calibration using the support set (Regression cal.), weighted k-nearest neighbor prediction from support examples (kNN), fine-tuning with the support set (FT) \cite{lee_few-shot_2023}, geometric order learning with weighted kNN prediction (GOL) \cite{lee_geometric_2022}.

\begin{table*}[h]
\centering
\caption{Comparison of methods on the DRIFT dataset (5-shot). Average $R^2$ score across 3 random runs for FT and 10 random runs for the rest. 
}
\adjustbox{max width=\textwidth}{\begin{tabular}{ccccc|ccc|c}
 \toprule
 \multicolumn{2}{c}{Source} & \multicolumn{3}{c}{Denmark} & \multicolumn{3}{c}{France} \\
  \multicolumn{2}{c}{Target} & Spain & Slovakia & Slovenia & Spain & Slovakia & Slovenia & \textit{avg}\\
 \midrule
\parbox[t]{2mm}{\multirow{7}{*}{\rotatebox[origin=c]{90}{Height}}} & Regression & 0.35 & -.05& 0.23 & 0.61 & 0.11 & 0.13 & 0.23 \\
&  Regression cal. & $0.17\scriptstyle{\pm0.12}$ & $0.49\scriptstyle{\pm0.10}$& $0.67\scriptstyle{\pm0.03}$ & $0.61\scriptstyle{\pm0.06}$ & $0.42\scriptstyle{\pm0.06}$ & $0.67\scriptstyle{\pm0.02}$ & 0.50\\ 
& FT \cite{lee_few-shot_2023} & $0.34\scriptstyle{\pm0.00}$ & $0.45\scriptstyle{\pm0.02}$ & $0.60\scriptstyle{\pm0.02}$ & $0.69\scriptstyle{\pm0.01}$ & $\mathbf{0.49\scriptstyle{\pm0.00}}$ & $0.50\scriptstyle{\pm0.09}$ & $0.51$\\
& kNN & $0.21\scriptstyle{\pm0.06}$ & $0.45\scriptstyle{\pm0.04}$& $0.68\scriptstyle{\pm0.03}$ & $0.55\scriptstyle{\pm0.04}$ & $0.38\scriptstyle{\pm0.06}$ & $0.63\scriptstyle{\pm0.03}$ & 0.48 \\ 
  & GOL \cite{lee_geometric_2022} & $0.21\scriptstyle{\pm0.12}$ & $0.35\scriptstyle{\pm0.13}$ & $0.67\scriptstyle{\pm0.04}$ & $0.62\scriptstyle{\pm0.05}$ & $0.41\scriptstyle{\pm0.08}$ & $0.65\scriptstyle{\pm0.03}$ & $0.49$\\
 & GOL+MDR & $\mathbf{0.60\scriptstyle{\pm0.04}}$ & $0.49\scriptstyle{\pm0.06}$ & $0.65\scriptstyle{\pm0.05}$ & $0.65\scriptstyle{\pm0.04}$ & $0.46\scriptstyle{\pm0.05}$ & $0.66\scriptstyle{\pm0.04}$ & $0.59$\\
 & GOL+FT+MDR & $0.56\scriptstyle{\pm0.07}$ & $\mathbf{0.57\scriptstyle{\pm0.03}}$ & $\mathbf{0.69\scriptstyle{\pm0.01}}$ & $\mathbf{0.70\scriptstyle{\pm0.01}}$ & $0.47\scriptstyle{\pm0.01}$ & $\mathbf{0.71\scriptstyle{\pm0.01}}$ & $\mathbf{0.62}$\\
\midrule
 \parbox[t]{2mm}{\multirow{7}{*}{\rotatebox[origin=c]{90}{Count}}} & Regression & $\mathbf{0.53}$ & 0.36 & 0.43 & 0.51 & 0.43 & 0.37 & 0.44 \\ 
& Regression cal. & $0.51\scriptstyle{\pm0.04}$ & $0.53\scriptstyle{\pm0.05}$& $0.59\scriptstyle{\pm0.03}$ & $0.53\scriptstyle{\pm0.07}$ & $0.61\scriptstyle{\pm0.04}$ & $\mathbf{0.58\scriptstyle{\pm0.04}}$ & $\mathbf{0.56}$\\ 
& FT \cite{lee_few-shot_2023} & $\mathbf{0.53\scriptstyle{\pm0.00}}$ & $0.48\scriptstyle{\pm0.00}$ & $\mathbf{0.60\scriptstyle{\pm0.01}}$ & $\mathbf{0.54\scriptstyle{\pm0.00}}$ & $0.53\scriptstyle{\pm0.01}$ & $0.40\scriptstyle{\pm0.00}$ & $0.51$\\
& kNN & $0.42\scriptstyle{\pm0.07}$ & $0.43\scriptstyle{\pm0.06}$& $0.55\scriptstyle{\pm0.03}$ & $0.51\scriptstyle{\pm0.06}$ & $0.55\scriptstyle{\pm0.14}$ & $0.55\scriptstyle{\pm0.08}$ & 0.50\\ 
& GOL \cite{lee_geometric_2022} & $0.51\scriptstyle{\pm0.03}$ & $\mathbf{0.56\scriptstyle{\pm0.03}}$ & $0.58\scriptstyle{\pm0.05}$ & $0.51\scriptstyle{\pm0.05}$ & $0.58\scriptstyle{\pm0.05}$ & $0.46\scriptstyle{\pm0.06}$ & $0.53$\\
& GOL+MDR & $0.51\scriptstyle{\pm0.04}$ & $0.55\scriptstyle{\pm0.06}$ & $0.59\scriptstyle{\pm0.03}$ & $\mathbf{0.54\scriptstyle{\pm0.04}}$ & $0.61\scriptstyle{\pm0.03}$ & $0.54\scriptstyle{\pm0.04}$ & $\mathbf{0.56}$ \\
& GOL+FT+MDR & $\mathbf{0.53\scriptstyle{\pm0.01}}$ & $0.49\scriptstyle{\pm0.10}$ & $0.59\scriptstyle{\pm0.05}$ & $0.51\scriptstyle{\pm0.09}$ & $\mathbf{0.63\scriptstyle{\pm0.01}}$ & $0.53\scriptstyle{\pm0.03}$ & $0.55$\\
  \midrule
\parbox[t]{2mm}{\multirow{7}{*}{\rotatebox[origin=c]{90}{Cover}}} & Regression & 0.11 & 0.70 & $\mathbf{0.81}$ & 0.80 & 0.77 & 0.76 & 0.66 \\ 
& Regression cal. & $0.41\scriptstyle{\pm0.04}$ & $0.75\scriptstyle{\pm0.01}$& $0.77\scriptstyle{\pm0.04}$ & $0.83\scriptstyle{\pm0.02}$ & $0.79\scriptstyle{\pm0.01}$ & $0.77\scriptstyle{\pm0.04}$ & $\mathbf{0.72}$\\ 
& FT \cite{lee_few-shot_2023} & $0.27\scriptstyle{\pm0.00}$ & $0.44\scriptstyle{\pm0.22}$ & $0.75\scriptstyle{\pm0.01}$ & $0.81\scriptstyle{\pm0.01}$ & $0.81\scriptstyle{\pm0.00}$ & $\mathbf{0.81\scriptstyle{\pm0.00}}$ & $0.65$\\
& kNN & $0.47\scriptstyle{\pm0.04}$ & $0.69\scriptstyle{\pm0.02}$& $0.71\scriptstyle{\pm0.06}$ & $0.75\scriptstyle{\pm0.04}$ & $0.73\scriptstyle{\pm0.05}$ & $0.68\scriptstyle{\pm0.07}$ & 0.67\\ 
& GOL \cite{lee_geometric_2022} & $0.51\scriptstyle{\pm0.06}$ & $\mathbf{0.76\scriptstyle{\pm0.02}}$ & $0.69\scriptstyle{\pm0.10}$ & $0.81\scriptstyle{\pm0.03}$ & $\mathbf{0.82\scriptstyle{\pm0.02}}$ & $0.71\scriptstyle{\pm0.09}$ & $\mathbf{0.72}$\\
 & GOL+MDR & $0.57\scriptstyle{\pm0.08}$ & $0.71\scriptstyle{\pm0.03}$ & $0.70\scriptstyle{\pm0.06}$ & $0.82\scriptstyle{\pm0.03}$ & $0.79\scriptstyle{\pm0.03}$ & $0.70\scriptstyle{\pm0.06}$ & $0.71$\\
& GOL+FT+MDR & $\mathbf{0.65\scriptstyle{\pm0.05}}$ & $0.74\scriptstyle{\pm0.04}$ & $0.64\scriptstyle{\pm0.05}$ & $\mathbf{0.84\scriptstyle{\pm0.01}}$ & $0.76\scriptstyle{\pm0.03}$ & $0.70\scriptstyle{\pm0.07}$ & $\mathbf{0.72}$ \\
 \bottomrule
  \end{tabular}}
  \label{tab:quantitative_fewshot}
\end{table*}

The off-the-shelf regressor has overall low and unstable performance, as expected. Calibration with the support set brings a significant but inconsistent boost in performance. This is particularly visible on the Denmark $\rightarrow$ Spain setup, for which calibration helps only on the cover task. Similarly, fine-tuning sometimes improves performance, but it relies on how well the underlying model generalizes to the target dataset. All inductive methods experience failure cases, with $R^2$ scores below $0.3$ in at least one setup.

\begin{figure}[h]
  \centering
  \includegraphics[width=0.85\linewidth]{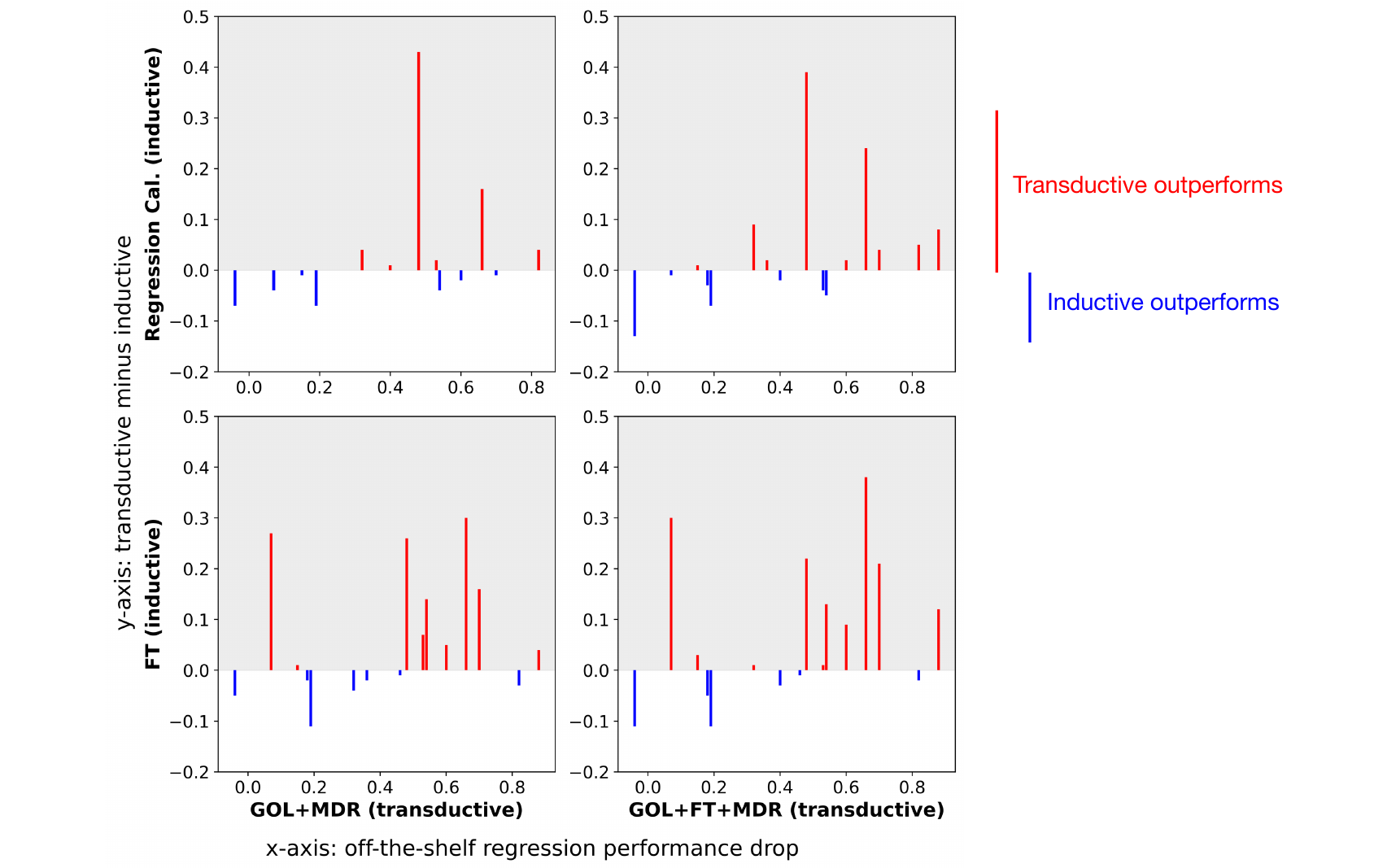}
  \caption{Transductive methods outperform inductive methods when the domain gap is larger. We plot pairwise performance differences between inductive and transductive methods on the y-axis, with off-the-shelf regression performance drop (source domain performance minus target domain performance) on the x-axis, an ad-hoc indicator of domain gap. 
  }
  \label{fig:transd_ind}
\end{figure}

Transductive approaches, on the other hand, decouple to some extent the performance on the target domain from the source domain. kNN performs overall more consistently than inductive approaches. It however performs worse than calibrated regression on average. GOL has a similar behavior, with a minor increase in performance undoubtedly due to better regularization in the source domain. GOL+MDR performs the best among transductive approaches, and especially well on the difficult setups of Denmark $\rightarrow$ Spain. Combining with fine-tuning on the support set further increases performance on some setups.

We note that the performance of direct cross-domain regression can serve as an indicator of a multifactorial domain gap. For example, on cover regression, where the label shift is limited, direct regression performs well, except for Denmark $\rightarrow$ Spain which is arguably the largest visual and semantic domain shift (Supplementary Fig. 2). We plot in Fig.~\ref{fig:transd_ind} a pairwise comparison of the two best-performing inductive and transductive methods against off-the-shelf regression performance drop. When the implied domain gap becomes large enough for inductive methods to fail, transductive methods are favored.

We repeat experiments with VGG16 \cite{simonyan_very_2014}, another popular feature extractor. GOL+MDR and GOL outperformed regression baselines for height and count with larger margins than using the ViT-B/16 backbone (Supplementary Table 4).   

\subsubsection{In-Domain Performance.} We report in Table \ref{tab:in-domain performance} the performance of GOL in the source domain. As reported in the original publication \cite{lee_geometric_2022}, GOL performs on par with ordinary regression.

\begin{table}[]
    \centering
    \caption{Source domain test performance. Average $R^2$ score across two source domains.}
    \begin{tabular}{ccccc}
    \toprule
        Backbone & Model & Height & Count & Cover \\
        \midrule
        \multirow{2}{*}{ViT-B/16} & GOL \cite{lee_geometric_2022} & $0.86\scriptstyle{\pm0.05}$ & $0.92\scriptstyle{\pm0.01}$ & $0.87\scriptstyle{\pm0.08}$\\
        \multicolumn{1}{l}{} & Regression & $0.88\scriptstyle{\pm0.05}$ & $0.90\scriptstyle{\pm0.01}$ & $0.86\scriptstyle{\pm0.09}$ \\
        \midrule
        \multirow{2}{*}{VGG16} & GOL \cite{lee_geometric_2022} & $0.80\scriptstyle{\pm0.07}$ & $0.91\scriptstyle{\pm0.01}$ & $0.80\scriptstyle{\pm0.13}$\\
        \multicolumn{1}{l}{} & Regression & $0.84\scriptstyle{\pm0.06}$ & $0.93\scriptstyle{\pm0.01}$ & $0.86\scriptstyle{\pm0.09}$ \\
        \bottomrule
    \end{tabular}
    \label{tab:in-domain performance}
\end{table}

\subsection{Ablation Studies}

We report in Table \ref{ablation} the effect of removing components of GOL+MDR, averaging across countries in the DRIFT dataset. MDR does not improve performance compared to the baseline when applied without ordered embeddings: the manifold in the target domain is too unstructured to give meaningful similarities. Ordered embeddings (GOL) generalize better in all tasks, and combining GOL+MDR gives the best results in 5 out of 6 experiments.

\begin{table*}[h]

\centering
\caption{Ablation study for combining GOL with the proposed MDR. Average $R^2$ score across two source domains and three target domains. We report the performance of a model with only the center loss and weighted kNN prediction as the baseline.
}
\begin{tabular}{cccccc}
 \toprule

Backbone & GOL &  MDR & Height & Count & Cover \\
\midrule
\multirow{4}{*}{ViT-B/16} & \checkmark & \checkmark & \multicolumn{1}{c}{$\mathbf{0.59\scriptstyle{\pm0.08}}$} & \multicolumn{1}{c}{$\mathbf{0.56\scriptstyle{\pm0.03}}$} & \multicolumn{1}{c}{$0.71\scriptstyle{\pm0.08}$} \\
\multicolumn{1}{l}{} & \checkmark &  & \multicolumn{1}{c}{$0.49\scriptstyle{\pm0.17}$} & \multicolumn{1}{c}{$0.53\scriptstyle{\pm0.04}$} & \multicolumn{1}{c}{$\mathbf{0.72\scriptstyle{\pm0.10}}$} \\
\multicolumn{1}{l}{} &  & \checkmark & \multicolumn{1}{c}{$0.40\scriptstyle{\pm0.17}$} & \multicolumn{1}{c}{$0.43\scriptstyle{\pm0.11}$} & \multicolumn{1}{c}{$0.65\scriptstyle{\pm0.10}$} \\
\multicolumn{1}{l}{} &  &  & \multicolumn{1}{c}{$0.48\scriptstyle{\pm0.16}$} & \multicolumn{1}{c}{$0.50\scriptstyle{\pm0.06}$} & \multicolumn{1}{c}{$0.67\scriptstyle{\pm0.09}$} \\
\midrule
\multirow{4}{*}{VGG16} & \checkmark & \checkmark & \multicolumn{1}{c}{$\mathbf{0.55\scriptstyle{\pm0.09}}$} & \multicolumn{1}{c}{$\mathbf{0.53\scriptstyle{\pm0.05}}$} & \multicolumn{1}{c}{$\mathbf{0.67\scriptstyle{\pm0.07}}$} \\
\multicolumn{1}{l}{} & \checkmark &  & \multicolumn{1}{c}{$0.47\scriptstyle{\pm0.16}$} & \multicolumn{1}{c}{$0.51\scriptstyle{\pm0.05}$} & \multicolumn{1}{c}{$0.64\scriptstyle{\pm0.09}$} \\
\multicolumn{1}{l}{} &  & \checkmark & \multicolumn{1}{c}{$0.25\scriptstyle{\pm0.16}$} & \multicolumn{1}{c}{$0.39\scriptstyle{\pm0.11}$} & \multicolumn{1}{c}{$0.59\scriptstyle{\pm0.08}$} \\
\multicolumn{1}{l}{} &  &  & \multicolumn{1}{c}{$0.31\scriptstyle{\pm0.13}$} & \multicolumn{1}{c}{$0.34\scriptstyle{\pm0.11}$} & \multicolumn{1}{c}{$0.61\scriptstyle{\pm0.08}$}  \\
 \bottomrule
  \end{tabular}
  \label{ablation}
\end{table*}

We test the sensitivity of different methods to the number of shots (Supplementary Fig. 5), i.e., the number of support samples per group. 

As often observed in few-shot setups, performance saturates after a certain number of shots \cite{nguyen_inductive_2022, snell_prototypical_2017}, here 20. Our proposed GOL+MDR outperforms the other methods on most of the setups, except for the easiest task of tree cover, where finetuning is needed to reach comparable performance as other baselines.

An important parameter in our approach is $k_v$, the number of support samples used for weighted sum prediction after MDR. We tested a set of values and observed a safe range around $k_v=2N$ (Supplementary Fig. 6).

\section{Conclusion}

We introduce the DRIFT dataset with forest-related image regression tasks across five European countries. We reveal different behaviors depending on tasks and target countries, which underlines the importance of taking into account the domain gap when choosing between inductive and transductive approaches. DRIFT covers a panel of situations with varying levels of difficulty and domain gaps. As such, it is close to real-world conditions, where those factors are difficult to estimate beforehand. In this context, we argue that transductive methods are a good starting point, as they allow qualitative analysis of the embedding space, even before having to label samples for quantitative evaluation. In contrast, inductive approaches remain "black boxes" with poor explainability.

We verify qualitatively and quantitatively that the embedding order can be maintained across domains, which unlocks advanced adaptation with diffusion approaches. This works particularly well in low-data regimes, and when the domain gap is high. 

\section*{Acknowledgements}

PC acknowledges support from the European Space Agency Climate Change Initiative (ESA-CCI) Biomass project (ESA ESRIN/ 4000123662) and RECCAP2 project 1190 (ESA ESRIN/ 4000123002/ 18/I-NB) and the ANR BMBF French German AI4FOREST (ANR-22-FAI1-0002 ).\\
MB and SL were supported by the European Research Council (ERC) under the European Union’s Horizon 2020 research and innovation programme (grant agreement no. 947757 TOFDRY) and a DFF Sapere Aude grant (no.~9064–00049B).\\
DG ackowledge support by the European Union’s Eurostars programme through the C-Trees project, grant number E114613.\\
XT acknowledges funding from the European Research Council (ERC) under the European Union’s Horizon 2020 research and innovation programme (grant agreement no. 947757 TOFDRY).\\

\noindent\textit{Image Credits.} Figures~\ref{fig:idea}, \ref{fig:challenge} and Supplementary Figures 1, 
3, and 4 contain data from the Danish Agency for Data Supply and Infrastructure (2018), the BD ORTHO© product from the French Institute for Geography (2018-2020), the Orthophoto and ALS products from the Slovakian authorities  GKÚ Bratislava and ÚGKK SR (2021), the PNOA product from the Spanish Institute of Geography (scne.es, 2020), the SkySat sensor from Planet labs PBC, and the Slovenian Environment Agency (ARSO, 2021).

\bibliographystyle{splncs04}


\begin{thebibliography}{10}
\providecommand{\url}[1]{\texttt{#1}}
\providecommand{\urlprefix}{URL }
\providecommand{\doi}[1]{https://doi.org/#1}

\bibitem{fao_tree_def}
Trees outside forests - towards a better awareness, \url{https://www.fao.org/3/y2328e/y2328e25.htm}

\bibitem{akiva_self-supervised_2022}
Akiva, P., Purri, M., Leotta, M.: Self-supervised material and texture representation learning for remote sensing tasks. In: Proceedings of the {IEEE}/{CVF} Conference on Computer Vision and Pattern Recognition ({CVPR}). pp. 8203--8215 (2022)

\bibitem{Beery_2018_ECCV}
Beery, S., Van~Horn, G., Perona, P.: Recognition in terra incognita. In: Proceedings of the European Conference on Computer Vision (ECCV). pp. 472--489 (2018)

\bibitem{beery_auto_2022}
Beery, S., Wu, G., Edwards, T., Pavetic, F., Majewski, B., Mukherjee, S., Chan, S., Morgan, J., Rathod, V., Huang, J.: The auto arborist dataset: A large-scale benchmark for multiview urban forest monitoring under domain shift. In: 2022 {IEEE}/{CVF} Conference on Computer Vision and Pattern Recognition ({CVPR}). pp. 21262--21275 (2022)

\bibitem{brandt_unexpectedly_2020}
Brandt, M., Tucker, C.J., Kariryaa, A., Rasmussen, K., Abel, C., Small, J., Chave, J., Rasmussen, L.V., Hiernaux, P., Diouf, A.A., Kergoat, L., Mertz, O., Igel, C., Gieseke, F., Schöning, J., Li, S., Melocik, K., Meyer, J., Sinno, S., Romero, E., Glennie, E., Montagu, A., Dendoncker, M., Fensholt, R.: An unexpectedly large count of trees in the west african sahara and sahel. Nature  \textbf{587}(7832),  78--82 (2020)

\bibitem{burgert_effects_2022}
Burgert, T., Ravanbakhsh, M., Demir, B.: On the effects of different types of label noise in multi-label remote sensing image classification. IEEE Transactions on Geoscience and Remote Sensing  (2022)

\bibitem{chen_transformer_2022}
Chen, G., Shang, Y.: Transformer for tree counting in aerial images. Remote Sensing  \textbf{14}(3), ~476 (2022)

\bibitem{chen_representation_2021}
Chen, X., Wang, S., Wang, J., Long, M.: Representation subspace distance for domain adaptation regression. In: International Conference on Machine Learning. pp. 1749--1759 (2021)

\bibitem{deng_comparison_2016}
Deng, S., Katoh, M., Yu, X., Hyyppä, J., Gao, T.: Comparison of tree species classifications at the individual tree level by combining {ALS} data and {RGB} images using different algorithms. Remote Sensing  \textbf{8}(12), ~1034 (2016)

\bibitem{donoser_diffusion_2013}
Donoser, M., Bischof, H.: Diffusion processes for retrieval revisited. In: 2013 {IEEE} Conference on Computer Vision and Pattern Recognition. pp. 1320--1327 (2013)

\bibitem{dosovitskiy_image_2021}
Dosovitskiy, A., Beyer, L., Kolesnikov, A., Weissenborn, D., Zhai, X., Unterthiner, T., Dehghani, M., Minderer, M., Heigold, G., Gelly, S., Uszkoreit, J., Houlsby, N.: An image is worth 16x16 words: Transformers for image recognition at scale. In: International Conference on Learning Representations (2021)

\bibitem{drivendata_biomassters_nodate}
{DrivenData}: The {BioMassters}, \url{https://www.drivendata.org/competitions/99/biomass- estimation/page/534/}

\bibitem{fanelli_random_2013}
Fanelli, G., Dantone, M., Gall, J., Fossati, A., Van~Gool, L.: Random {Forests} for {Real} {Time} {3D} {Face} {Analysis}. International Journal of Computer Vision  \textbf{101}(3),  437--458 (2013)

\bibitem{fayad_vision_2023}
Fayad, I., Ciais, P., Schwartz, M., Wigneron, J.P., Baghdadi, N., {de Truchis}, A., d'Aspremont, A., Frappart, F., Saatchi, S., Sean, E., Pellissier-Tanon, A., Bazzi, H.: Hy-tec: a hybrid vision transformer model for high-resolution and large-scale mapping of canopy height. Remote Sensing of Environment  \textbf{302},  113945 (2024)

\bibitem{feng_doubling_2022}
Feng, Y., Zeng, Z., Searchinger, T.D., Ziegler, A.D., Wu, J., Wang, D., He, X., Elsen, P.R., Ciais, P., Xu, R., Guo, Z., Peng, L., Tao, Y., Spracklen, D.V., Holden, J., Liu, X., Zheng, Y., Xu, P., Chen, J., Jiang, X., Song, X.P., Lakshmi, V., Wood, E.F., Zheng, C.: Doubling of annual forest carbon loss over the tropics during the early twenty-first century. Nature Sustainability  \textbf{5}(5),  444--451 (2022)

\bibitem{finn_model-agnostic_2017}
Finn, C., Abbeel, P., Levine, S.: Model-agnostic meta-learning for fast adaptation of deep networks. In: Proceedings of the 34th International Conference on Machine Learning. Proceedings of Machine Learning Research, vol.~70, pp. 1126--1135. {PMLR} (2017)

\bibitem{girard_noisy_2019}
Girard, N., Charpiat, G., Tarabalka, Y.: Noisy supervision for correcting misaligned cadaster maps without perfect ground truth data. In: 2019 {IEEE} International Geoscience and Remote Sensing Symposium. pp. 10103--10106 (2019)

\bibitem{hosu2024uhd}
Hosu, V., Agnolucci, L., Wiedemann, O., Iso, D.: Uhd-iqa benchmark database: Pushing the boundaries of blind photo quality assessment (2024), arXiv preprint arXiv:2406.17472

\bibitem{hudak_carbon_2020}
Hudak, A.T., Fekety, P.A., Kane, V.R., Kennedy, R.E., Filippelli, S.K., Falkowski, M.J., Tinkham, W.T., Smith, A.M.S., Crookston, N.L., Domke, G.M., Corrao, M.V., Bright, B.C., Churchill, D.J., Gould, P.J., McGaughey, R.J., Kane, J.T., Dong, J.: A carbon monitoring system for mapping regional, annual aboveground biomass across the northwestern {USA}. Environmental Research Letters  \textbf{15}(9),  095003 (2020)

\bibitem{iscen_label_2019}
Iscen, A., Tolias, G., Avrithis, Y., Chum, O.: Label propagation for deep semi-supervised learning. In: Proceedings of the {IEEE}/{CVF} Conference on Computer Vision and Pattern Recognition ({CVPR}) (2019)

\bibitem{iscen_efficient_2017}
Iscen, A., Tolias, G., Avrithis, Y., Furon, T., Chum, O.: Efficient diffusion on region manifolds: Recovering small objects with compact {CNN} representations. In: 2017 {IEEE} Conference on Computer Vision and Pattern Recognition ({CVPR}). pp. 926--935 (2017)

\bibitem{jucker_allometric_2017}
Jucker, T., Caspersen, J., Chave, J., Antin, C., Barbier, N., Bongers, F., Dalponte, M., van Ewijk, K.Y., Forrester, D.I., Haeni, M., Higgins, S.I., Holdaway, R.J., Iida, Y., Lorimer, C., Marshall, P.L., Momo, S., Moncrieff, G.R., Ploton, P., Poorter, L., Rahman, K.A., Schlund, M., Sonké, B., Sterck, F.J., Trugman, A.T., Usoltsev, V.A., Vanderwel, M.C., Waldner, P., Wedeux, B.M.M., Wirth, C., Wöll, H., Woods, M., Xiang, W., Zimmermann, N.E., Coomes, D.A.: Allometric equations for integrating remote sensing imagery into forest monitoring programmes. Global Change Biology  \textbf{23}(1),  177--190 (2017)

\bibitem{kalinicheva_multi-layer_2022}
Kalinicheva, E., Landrieu, L., Mallet, C., Chehata, N.: Multi-layer modeling of dense vegetation from aerial {LiDAR} scans. In: Proceedings of the {IEEE}/{CVF} Conference on Computer Vision and Pattern Recognition ({CVPR}) Workshops. pp. 1342--1351 (2022)

\bibitem{kattenborn_convolutional_2020}
Kattenborn, T., Eichel, J., Wiser, S., Burrows, L., Fassnacht, F.E., Schmidtlein, S.: Convolutional {Neural} {Networks} accurately predict cover fractions of plant species and communities in {Unmanned} {Aerial} {Vehicle} imagery. Remote Sensing in Ecology and Conservation  \textbf{6}(4),  472--486 (2020)

\bibitem{knapp_linking_2018}
Knapp, N., Fischer, R., Huth, A.: Linking lidar and forest modeling to assess biomass estimation across scales and disturbance states. Remote Sensing of Environment  \textbf{205},  199--209 (2018)

\bibitem{kundu_universal_2020}
Kundu, J.N., Venkat, N., M~V, R., Babu, R.V.: Universal source-free domain adaptation. In: Proceedings of the {IEEE}/{CVF} Conference on Computer Vision and Pattern Recognition. (2020)

\bibitem{kuprashevich2023mivolo}
Kuprashevich, M., Tolstykh, I.: Mivolo: Multi-input transformer for age and gender estimation. In: International Conference on Analysis of Images, Social Networks and Texts. pp. 212--226. Springer (2023)

\bibitem{lang_high-resolution_2023}
Lang, N., Jetz, W., Schindler, K., Wegner, J.D.: A high-resolution canopy height model of the {Earth}. Nature Ecology \& Evolution  \textbf{7}(11),  1778--1789 (2023)

\bibitem{lee_order_2022}
Lee, S.H., Kim, C.S.: Order learning using partially ordered data via chainization. In: {ECCV} 2022. pp. 196--211 (2022)

\bibitem{lee_geometric_2022}
Lee, S.H., Shin, N.H., Kim, C.S.: Geometric order learning for rank estimation. In: Advances in Neural Information Processing Systems. vol.~35, pp. 27--39 (2022)

\bibitem{lee_few-shot_2023}
Lee, S., Seo, S., Kim, J., Lee, Y., Hwang, S.: Few-shot fine-tuning is all you need for source-free domain adaptation (2023), arXiv preprint arXiv:2304.00792

\bibitem{7301352}
Levi, G., Hassncer, T.: Age and gender classification using convolutional neural networks. In: 2015 IEEE Conference on Computer Vision and Pattern Recognition Workshops (CVPRW). pp. 34--42 (2015)

\bibitem{li_deep_2023}
Li, S., Brandt, M., Fensholt, R., Kariryaa, A., Igel, C., Gieseke, F., Nord-Larsen, T., Oehmcke, S., Carlsen, A.H., Junttila, S., Tong, X., d’Aspremont, A., Ciais, P.: Deep learning enables image-based tree counting, crown segmentation and height prediction at national scale. PNAS Nexus  \textbf{2}(4) (2023)

\bibitem{li2023deep}
Li, S., Brandt, M., Fensholt, R., Kariryaa, A., Igel, C., Gieseke, F., Nord-Larsen, T., Oehmcke, S., Carlsen, A.H., Junttila, S., et~al.: Deep learning enables image-based tree counting, crown segmentation, and height prediction at national scale. PNAS nexus  \textbf{2}(4) (2023)

\bibitem{li2023deepbiomass}
Li, S., Brandt, M., Tong, X., Oehmcke, S., Igel, C., Gieseke, F., Nord-Larsen, T., Fensholt, R., Chave, J., Ciais, P.: Deep learning tree and forest biomass from sub-meter resolution images (2023), researchSquare preprint 10.21203/rs.3.rs-3335298

\bibitem{li_rse_2020}
Li, W., Buitenwerf, R., Munk, M., Bøcher, P.K., Svenning, J.C.: Deep-learning based high-resolution mapping shows woody vegetation densification in greater maasai mara ecosystem. Remote Sensing of Environment  \textbf{247},  111953 (2020)

\bibitem{li_forest_2020}
Li, Y., Li, M., Li, C., Liu, Z.: Forest aboveground biomass estimation using landsat 8 and sentinel-1a data with machine learning algorithms. Scientific Reports  \textbf{10}(1), ~9952 (2020)

\bibitem{liang_we_2020}
Liang, J., Hu, D., Feng, J.: Do we really need to access the source data? source hypothesis transfer for unsupervised domain adaptation. In: International Conference on Machine Learning ({ICML}). pp. 6028--6039 (2020)

\bibitem{lim_order_2020}
Lim, K., Shin, N.H., Lee, Y.Y., Kim, C.S.: Order learning and its application to age estimation. In: International Conference on Learning Representations (2020)

\bibitem{lin2021fpage}
Lin, Y., Shen, J., Wang, Y., Pantic, M.: Fp-age: Leveraging face parsing attention for facial age estimation in the wild (2021), arXiv preprint arXiv:2106.11145

\bibitem{liu_overlooked_2023}
Liu, S., Brandt, M., Nord-Larsen, T., Chave, J., Reiner, F., Lang, N., Tong, X., Ciais, P., Igel, C., Pascual, A., Guerra-Hernandez, J., Li, S., Mugabowindekwe, M., Saatchi, S., Yue, Y., Chen, Z., Fensholt, R.: The overlooked contribution of trees outside forests to tree cover and woody biomass across europe. Science Advances  \textbf{9}(37) (2023)

\bibitem{Marsocci_2023_CVPR}
Marsocci, V., Gonthier, N., Garioud, A., Scardapane, S., Mallet, C.: Geomultitasknet: Remote sensing unsupervised domain adaptation using geographical coordinates. In: Proceedings of the IEEE/CVF Conference on Computer Vision and Pattern Recognition (CVPR) Workshops. pp. 2075--2085 (2023)

\bibitem{mathelin_adversarial_2021}
Mathelin, A.d., Richard, G., Deheeger, F., Mougeot, M., Vayatis, N.: Adversarial weighting for domain adaptation in regression. In: 2021 {IEEE} 33rd International Conference on Tools with Artificial Intelligence ({ICTAI}). pp. 49--56. {IEEE} Computer Society (2021)

\bibitem{moschoglou2017agedb}
Moschoglou, S., Papaioannou, A., Sagonas, C., Deng, J., Kotsia, I., Zafeiriou, S.: Agedb: the first manually collected, in-the-wild age database. In: Proceedings of the IEEE Conference on Computer Vision and Pattern Recognition Workshop. vol.~2, p.~5 (2017)

\bibitem{mugabowindekwe_nation-wide_2023}
Mugabowindekwe, M., Brandt, M., Chave, J., Reiner, F., Skole, D.L., Kariryaa, A., Igel, C., Hiernaux, P., Ciais, P., Mertz, O., Tong, X., Li, S., Rwanyiziri, G., Dushimiyimana, T., Ndoli, A., Uwizeyimana, V., Lillesø, J.P.B., Gieseke, F., Tucker, C.J., Saatchi, S., Fensholt, R.: Nation-wide mapping of tree-level aboveground carbon stocks in rwanda. Nature Climate Change  \textbf{13}(1),  91--97 (2023)

\bibitem{6247954}
Murray, N., Marchesotti, L., Perronnin, F.: Ava: A large-scale database for aesthetic visual analysis. In: 2012 IEEE Conference on Computer Vision and Pattern Recognition. pp. 2408--2415 (2012)

\bibitem{nejjar_dare-gram_2023}
Nejjar, I., Wang, Q., Fink, O.: {DARE}-{GRAM} : Unsupervised domain adaptation regression by aligning inversed gram matrices. In: Proceedings of the {IEEE}/{CVF} Conference on Computer Vision and Pattern Recognition. (2023)

\bibitem{nguyen_inductive_2022}
Nguyen, K.D., Tran, Q.H., Nguyen, K., Hua, B.S., Nguyen, R.: Inductive and transductive few-shot video classification via appearance and temporal alignments. In: {ECCV} 2022. pp. 471--487 (2022)

\bibitem{pardoe_boosting_2010}
Pardoe, D., Stone, P.: Boosting for regression transfer. In: International Conference on Machine Learning (2010)

\bibitem{potapov_annual_2019}
Potapov, P., Tyukavina, A., Turubanova, S., Talero, Y., Hernandez-Serna, A., Hansen, M.C., Saah, D., Tenneson, K., Poortinga, A., Aekakkararungroj, A., Chishtie, F., Towashiraporn, P., Bhandari, B., Aung, K.S., Nguyen, Q.H.: Annual continuous fields of woody vegetation structure in the {Lower} {Mekong} region from 2000‐2017 {Landsat} time-series. Remote Sensing of Environment  \textbf{232},  111278 (2019)

\bibitem{1613043}
Ricanek, K., Tesafaye, T.: Morph: a longitudinal image database of normal adult age-progression. In: 7th International Conference on Automatic Face and Gesture Recognition (FGR06). pp. 341--345 (2006)

\bibitem{robin_learning_2022}
Robin, C., Requena-Mesa, C., Benson, V., Poehls, J., Alonzo, L., Carvalhais, N., Reichstein, M.: Learning to forecast vegetation greenness at fine resolution over {Africa} with {ConvLSTMs}. In: {NeurIPS} 2022 {Workshop} on {Tackling} {Climate} {Change} with {Machine} {Learning} (2022)

\bibitem{robinson_large_2019}
Robinson, C., Hou, L., Malkin, K., Soobitsky, R., Czawlytko, J., Dilkina, B., Jojic, N.: Large scale high-resolution land cover mapping with multi-resolution data. In: 2019 {IEEE}/{CVF} Conference on Computer Vision and Pattern Recognition ({CVPR}). pp. 12718--12727 (2019)

\bibitem{roussel_lidr_2020}
Roussel, J.R., Auty, D., Coops, N.C., Tompalski, P., Goodbody, T.R.H., Meador, A.S., Bourdon, J.F., de~Boissieu, F., Achim, A.: {lidR}: An r package for analysis of airborne laser scanning ({ALS}) data. Remote Sensing of Environment  \textbf{251},  112061 (2020)

\bibitem{schifanella2015imageworththousandfavorites}
Schifanella, R., Redi, M., Aiello, L.: An image is worth more than a thousand favorites: Surfacing the hidden beauty of flickr pictures (2015), arXiv preprint arXiv:1505.03358

\bibitem{shin_moving_2022}
Shin, N.H., Lee, S.H., Kim, C.S.: Moving window regression: a novel approach to ordinal regression. In: Proceedings of the {IEEE}/{CVF} conference on computer vision and pattern recognition (2022)

\bibitem{simonyan_very_2014}
Simonyan, K., Zisserman, A.: Very deep convolutional networks for large-scale image recognition. 3rd International Conference on Learning Representations (ICLR 2015) pp. 1--14 (2015)

\bibitem{snell_prototypical_2017}
Snell, J., Swersky, K., Zemel, R.: Prototypical networks for few-shot learning. In: Advances in Neural Information Processing Systems (2017)

\bibitem{sumbul_label_2023}
Sumbul, G., Demir, B.: Label noise robust image representation learning based on supervised variational autoencoders in remote sensing. In: {IEEE} International Geoscience and Remote Sensing Symposium ({IGARSS}) (2023)

\bibitem{teshima_few-shot_2020}
Teshima, T., Sato, I., Sugiyama, M.: Few-shot domain adaptation by causal mechanism transfer. In: Proceedings of the 37th International Conference on Machine Learning (2020)

\bibitem{Voulgaris_2023_CVPR}
Voulgaris, G., Philippides, A., Dolley, J., Reffin, J., Marshall, F., Quadrianto, N.: Seasonal domain shift in the global south: Dataset and deep features analysis. In: Proceedings of the IEEE/CVF Conference on Computer Vision and Pattern Recognition (CVPR) Workshops. pp. 2116--2124 (2023)

\bibitem{wang_transfer_2019}
Wang, B., Mendez, J., Cai, M., Eaton, E.: Transfer learning via minimizing the performance gap between domains. In: Advances in Neural Information Processing Systems. vol.~32. Curran Associates, Inc. (2019)

\bibitem{wang2024trustworthy}
Wang, S., Han, W., Huang, X., Zhang, X., Wang, L., Li, J.: Trustworthy remote sensing interpretation: Concepts, technologies, and applications. ISPRS Journal of Photogrammetry and Remote Sensing  \textbf{209},  150--172 (2024)

\bibitem{weinstein_benchmark_2021}
Weinstein, B.G., Graves, S.J., Marconi, S., Singh, A., Zare, A., Stewart, D., Bohlman, S.A., White, E.P.: A benchmark dataset for canopy crown detection and delineation in co-registered airborne {RGB}, {LiDAR} and hyperspectral imagery from the national ecological observation network. PLoS computational biology  \textbf{17}(7),  e1009180 (2021)

\bibitem{huang2016unsupervised}
Zhang, Y., Liu, L., Li, C., Loy, C.C.: Quantifying facial age by posterior of age comparisons. In: British Machine Vision Conference (BMVC) (2017)

\bibitem{zhang2017age}
Zhang, Z., Song, Y., Qi, H.: Age progression/regression by conditional adversarial autoencoder. In: Proceedings of the IEEE conference on computer vision and pattern recognition. pp. 5810--5818 (2017)

\bibitem{zhou_learning_2003}
Zhou, D., Bousquet, O., Lal, T., Weston, J., Schölkopf, B.: Learning with local and global consistency. In: Advances in Neural Information Processing Systems. vol.~16. {MIT} Press (2003)

\end{thebibliography}

\newpage

\extendeddatatab
\extendeddatafigs

\section*{\textit{Supplementary Materials}}
\section{DRIFT dataset details}

\textit{Tree counts.} For Denmark, the subset was built from the predictions of \cite{li_deep_2023}. Due to the relatively high reported counting performance, we used the image-level counts directly as count annotations. For the rest, we corrected CHM-derived tree counts using linear regression with 3-fold validation. Correction factors ($y=ax+b$) and final $R^2$ scores are reported in Supplementary Table \ref{tab:dataset_patch_counts}. \\

\begin{table}[h]
    \centering
     \caption{Image-level tree count annotation accuracy.}
    \begin{tabular}{c|c|c|c|c}
    \toprule
        Dataset & \# Images & Base $R^2$ & 3-fold $R^2$ with correction & Correction factors (b, a) \\
        \hline
        Denmark & 13094 & $0.93$ \cite{li_deep_2023} & - & -\\
        France & 10298 & $0.83$ & $0.81 \scriptstyle{\pm0.04}$ & $1.93$, $0.77$ \\
        Slovakia & 3129 & $0.85$ & $0.89 \scriptstyle{\pm0.03}$ & $1.79$, $0.77$ \\
        Slovenia & 1903 & $0.65$ & $0.77 \scriptstyle{\pm0.06}$ & $9.24$, $0.56$ \\
        Spain & 1150 & $0.87$ & $0.86 \scriptstyle{\pm0.04}$ & $10.5$, $0.79$ \\
        \bottomrule
    \end{tabular}
    \label{tab:dataset_patch_counts}
\end{table}

\noindent\textit{Label distribution.} For the Denmark subset, we picked images to have a roughly uniform distribution of canopy height due to a limited variation of height observed in our initial random image selection. For the France and Slovakia subsets, images were randomly picked without any filtering or visual inspection. For the Spain and Slovenia subsets, we purposely chose forest-dominated regions prior to random picking to ensure efficient coverage of forest landscapes. The exact sampling locations are indicated in Supplementary Fig.~\ref{fig:biom}. As such, the DRIFT dataset covers diverse label distributions representing different real-world scenarios.  \\

\noindent\textit{Labeling uncertainties.} The raw LiDAR point clouds have a horizontal (position) error of $\textsc{std}<20\text{cm}$ and a vertical (height) error of $\textsc{std}<10\text{cm}$, which are mitigated when extracting patch-level aggregates. We directly extracted tree cover fraction and non-zero mean height, without additional uncertainty. \\

\noindent\textit{Comparison with existing vegetation regression datasets.} We compare the DRIFT dataset to existing regression datasets for vegetation monitoring in Supplementary Table~\ref{tab:otherdatasets}. None of these datasets relies on sub-meter imagery over an area large enough to include significant domain shifts. Additionally, DRIFT includes three target variables with different levels of semantic complexity. \\

\begin{table}[h]
\centering
\caption{Vegetation regression datasets.}
\begin{tabular}{c|c|c|c|c}
       & Size & Target variable & Resolution & Domain shift\\
       \hline
      Potapov et al. 2019 \cite{potapov_annual_2019} & 56k & Cover, Height & 30m & No \\
      Kattenborn et al. 2020 \cite{kattenborn_convolutional_2020} & 4k & Cover & 5cm & No \\
      Hudak et al. 2020 \cite{hudak_carbon_2020} & 3.8k & Biomass & 30m & No\\
      Feng et al. 2022 \cite{feng_doubling_2022} & 24k & Cover, Age & 30m & Yes \\
      Robin et al. 2022 \cite{robin_learning_2022} & 40k & Greenness & 30m & Yes \\
      Lang et al. 2022 \cite{lang_high-resolution_2023} & 68k & Height & 10m & Yes\\
\end{tabular}
\label{tab:otherdatasets}
\end{table}

\FloatBarrier

\noindent \textit{Comparison with existing image-level regression datasets.} Image-level regression examples can be found in facial age estimation \cite{7301352}, aesthetical quality estimation \cite{schifanella2015imageworththousandfavorites}, facial pose estimation \cite{fanelli_random_2013}. There has been limited interest in cross-domain regression, however. We summarize popular benchmark datasets on image-level regression in Supplementary Table~\ref{tab:otherdatasets2}.\\ 

\begin{table}[h]
\centering
\caption{Image-level regression datasets in broader computer vision.}
\begin{tabular}{c|c|c|c}
       & Size & Target variable &  Domain shift\\
       \hline
       Zhang et al. 2017~\cite{zhang2017age}& 20k & Facial age & No \\
       Moschoglou et al. 2017~\cite{moschoglou2017agedb} & 16k& Facial age & No \\
       Huang et al. 2016~\cite{huang2016unsupervised}& 41k & Facial age & No \\
       Ricanek et al. 2006~\cite{1613043} & 55k & Facial age & No \\
       Lin et al. 2021~\cite{lin2021fpage}& 287k & Facial age & No \\
       Kuprashevich et al. 2023~\cite{kuprashevich2023mivolo}&67k & Facial age & No\\
       Hosu et al. 2024~\cite{hosu2024uhd} & 6k & Image quality & No \\
       Murray et al. 2012~\cite{6247954} & 255k & Aesthetic score & No\\
\end{tabular}
\label{tab:otherdatasets2}
\end{table}

\newpage

\noindent\textit{Data licenses}

\begin{enumerate}
    \item \textbf{Denmark}
    
    20cm ground resolution RGB aerial imagery from the Danish Agency for Data Supply and Infrastructure (SDFI), 2018. CHMs are from the Danmarks Højdemodel product by SDFI, CC BY 4.0 license.
    
    \item \textbf{France}
    
    20cm ground resolution RGB aerial imagery from the French Institute of Geography (IGN), BD ORTHO© product, 2018-2020. CHMs derived from the LiDAR HD product. All Data is under the 2.0 Open License.

    \item \textbf{Slovakia}
    
    20cm ground resolution RGB aerial imagery from GKÚ Bratislava, NLC, 2021, downloaded from the GeoPortal. CHMs derived from the national ALS campaign, credits to ÚGKK SR, CC BY 4.0 license.
    
    \item \textbf{Spain}
    
    25cm ground resolution RGB aerial imagery and CHMs derived from the Spanish Institute of Geography (IGN), PNOA 2020, scne.es. Data is under the CC-BY 4.0 License.

    \item \textbf{Slovenia}

    50cm ground resolution RGB satellite imagery bought from Planet, SkySat sensor, 2021. Any usage must be solely for noncommercial education or scientific research purposes, and publication in academic or scientific research journals. All such publications must include an attribution that clearly and conspicuously identifies Planet Labs PBC. CHMs derived from the national ALS campaign, open license with attribution: Ministry of the Environment and Spatial Planning, Slovenian Environment Agency (ARSO).

\end{enumerate}

\begin{figure}[h]
  \centering
\includegraphics[width=0.65\textwidth,keepaspectratio]{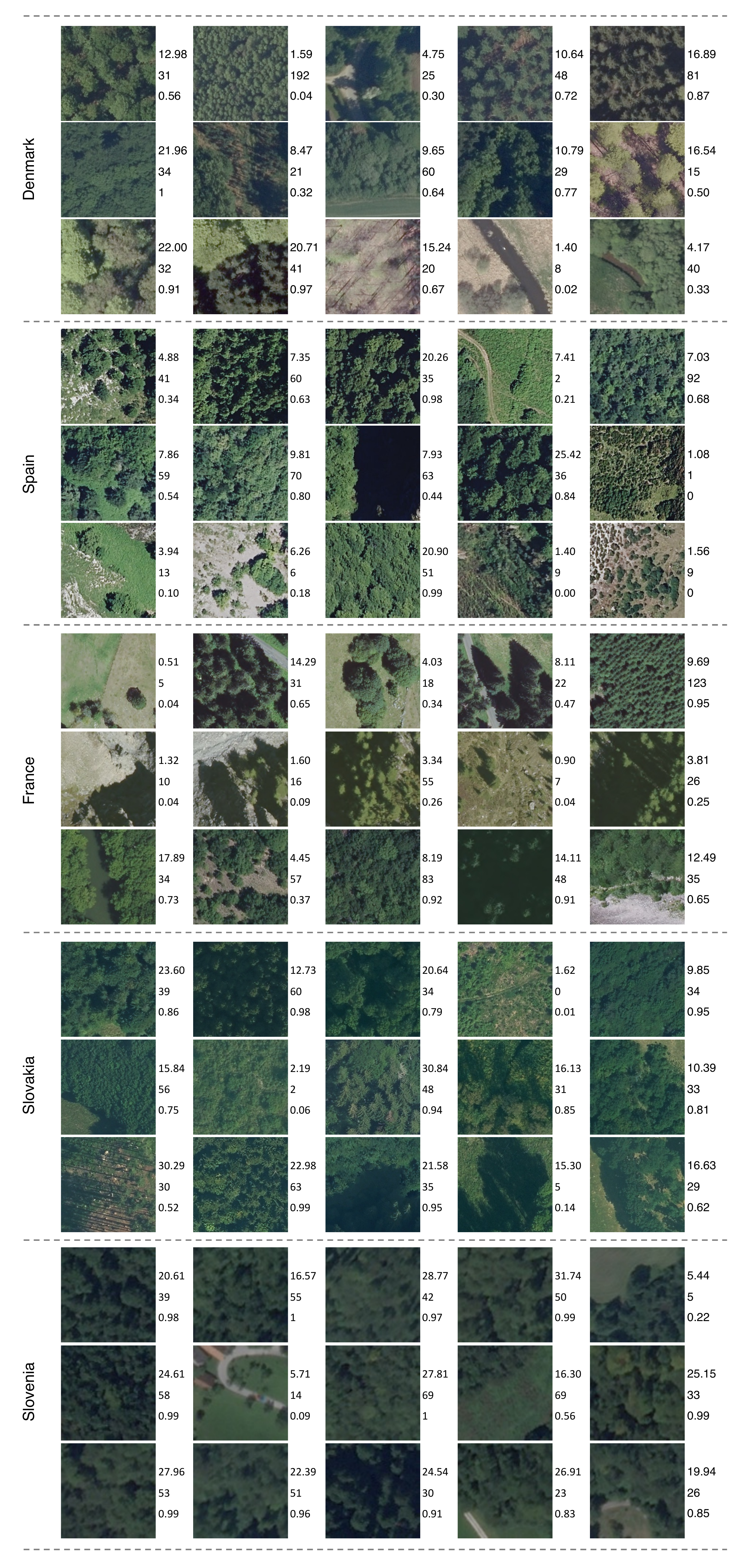}
  \caption{Examples in the DRIFT dataset. Labels on the right: canopy height in meters (1st row), tree count (2nd row), and tree cover fraction (3rd row).
  }
  \label{fig:data_example}
\end{figure}

\FloatBarrier

\section{VGG experiments}

We repeat experiments with VGG \cite{simonyan_very_2014} and report results in Supplementary Table~\ref{tab:quantitative_vgg}. Our conclusions remain unchanged, with GOL+MDR and GOL+FT+MDR performing overall better than other methods except in setups with good off-the-shelf inductive performance. Compared to ViT, performance with VGG as a feature extractor is significantly lower everywhere, but the drop is smaller for transductive methods. For example, GOL+MDR performance drops by $0.03$ in $R^2$ on average on counts, whereas FT performance drops by $0.27$. It is also interesting to note that the drop in performance is the smallest across methods for the cover task, which corroborates our intuition that it is an easier task.

\begin{table*}[h]

\centering
\caption{Comparison of methods on the DRIFT dataset (5-shot) with the VGG16 feature extractor. Average $R^2$ score across 3 random runs for FT and 10 random runs for the rest. 
}
\begin{tabular}{ccccc|ccc|c}
 \toprule
 \multicolumn{2}{c}{Source} & \multicolumn{3}{c}{Denmark} & \multicolumn{3}{c}{France} \\
  \multicolumn{2}{c}{Target} & Spain & Slovakia & Slovenia & Spain & Slovakia & Slovenia & \textit{avg}\\
 \midrule
\parbox[t]{2mm}{\multirow{7}{*}{\rotatebox[origin=c]{90}{Height}}} & Regression & 0.28 & 0.11 & -0.01 & 0.53 & -0.44 & -0.13 & 0.06 \\ 
& Regression cal. & $0.23\scriptstyle{\pm0.09}$ & $\mathbf{0.51\scriptstyle{\pm0.05}}$ & $0.41\scriptstyle{\pm0.04}$ & $0.55\scriptstyle{\pm0.05}$ & $0.34\scriptstyle{\pm0.08}$ & $0.35\scriptstyle{\pm0.02}$ & 0.40\\ 
& FT \cite{lee_few-shot_2023} & $0.20\scriptstyle{\pm0.06}$ & $0.40\scriptstyle{\pm0.01}$ & -$0.02\scriptstyle{\pm0.09}$ & $0.55\scriptstyle{\pm0.00}$ & -$0.03\scriptstyle{\pm0.12}$ & -$0.31\scriptstyle{\pm0.02}$ & 0.13\\
& kNN & $0.19\scriptstyle{\pm0.06}$ & $0.49\scriptstyle{\pm0.05}$& $0.39\scriptstyle{\pm0.10}$ & $0.41\scriptstyle{\pm0.04}$ & $0.17\scriptstyle{\pm0.07}$ & $0.20\scriptstyle{\pm0.09}$ & 0.31 \\ 
& GOL \cite{lee_geometric_2022} & $0.27\scriptstyle{\pm0.07}$ & $0.34\scriptstyle{\pm0.08}$ & $0.63\scriptstyle{\pm0.03}$ & $\mathbf{0.61\scriptstyle{\pm0.03}}$ & $0.34\scriptstyle{\pm0.04}$ & $0.64\scriptstyle{\pm0.04}$ & 0.47\\
& GOL+MDR & $0.58\scriptstyle{\pm0.03}$ & $0.43\scriptstyle{\pm0.04}$ & $0.62\scriptstyle{\pm0.06}$ & $0.58\scriptstyle{\pm0.04}$ & $0.41\scriptstyle{\pm0.03}$ & $\mathbf{0.66\scriptstyle{\pm0.02}}$ & 0.55\\
& GOL+FT+MDR & $\mathbf{0.59\scriptstyle{\pm0.01}}$ & $0.50\scriptstyle{\pm0.01}$ & $\mathbf{0.64\scriptstyle{\pm0.03}}$ & $0.54\scriptstyle{\pm0.03}$ & $\mathbf{0.44\scriptstyle{\pm0.02}}$ & $0.64\scriptstyle{\pm0.03}$ & $\mathbf{0.56}$ \\
\midrule
 \parbox[t]{2mm}{\multirow{7}{*}{\rotatebox[origin=c]{90}{Count}}} & Regression & 0.51 & 0.41 & -0.05 & 0.49 & 0.43 & -1.92 & -0.02 \\ 
& Regression cal. & $0.45\scriptstyle{\pm0.06}$ & $0.40\scriptstyle{\pm0.08}$ & $0.08\scriptstyle{\pm0.06}$ & $0.50\scriptstyle{\pm0.05}$ & $0.51\scriptstyle{\pm0.04}$ & $0.41\scriptstyle{\pm0.06}$ & 0.39\\ 
& FT \cite{lee_few-shot_2023} & $\mathbf{0.59\scriptstyle{\pm0.00}}$ & $0.45\scriptstyle{\pm0.02}$ & $0.19\scriptstyle{\pm0.02}$ & $0.49\scriptstyle{\pm0.01}$ & $0.52\scriptstyle{\pm0.02}$ & -$0.80\scriptstyle{\pm0.07}$ & 0.24 \\
& kNN & $0.40\scriptstyle{\pm0.08}$ & $0.45\scriptstyle{\pm0.06}$& $0.10\scriptstyle{\pm0.15}$ & $0.37\scriptstyle{\pm0.06}$ & $0.34\scriptstyle{\pm0.08}$ & $0.39\scriptstyle{\pm0.04}$ & 0.34\\ 
& GOL \cite{lee_geometric_2022} & $0.48\scriptstyle{\pm0.05}$ & $0.52\scriptstyle{\pm0.06}$ & $0.54\scriptstyle{\pm0.07}$ & $0.48\scriptstyle{\pm0.05}$ & $0.59\scriptstyle{\pm0.04}$ & $0.44\scriptstyle{\pm0.05}$ & 0.51\\
& GOL+MDR & $0.48\scriptstyle{\pm0.04}$ & $\mathbf{0.55\scriptstyle{\pm0.04}}$ & $\mathbf{0.58\scriptstyle{\pm0.04}}$ & $\mathbf{0.52\scriptstyle{\pm0.04}}$ & $\mathbf{0.60\scriptstyle{\pm0.03}}$ & $\mathbf{0.45\scriptstyle{\pm0.07}}$ & $\mathbf{0.53}$\\
& GOL+FT+MDR & $0.54\scriptstyle{\pm0.05}$ & $0.53\scriptstyle{\pm0.01}$ & $0.43\scriptstyle{\pm0.07}$ & $0.51\scriptstyle{\pm0.01}$ & $0.59\scriptstyle{\pm0.01}$ & $0.34\scriptstyle{\pm0.06}$ & 0.49\\
  \midrule
\parbox[t]{2mm}{\multirow{7}{*}{\rotatebox[origin=c]{90}{Cover}}} & Regression & 0.44 & 0.69 & 0.66 & 0.78 & 0.61 & 0.54 & 0.62\\ 
& Regression cal. & $0.61\scriptstyle{\pm0.02}$ & $\mathbf{0.79\scriptstyle{\pm0.01}}$ & $0.63\scriptstyle{\pm0.04}$ & $0.80\scriptstyle{\pm0.02}$ & $\mathbf{0.74\scriptstyle{\pm0.01}}$ & $0.62\scriptstyle{\pm0.06}$ & 0.70\\
& FT \cite{lee_few-shot_2023} & $0.40\scriptstyle{\pm0.02}$ & $0.67\scriptstyle{\pm0.03}$ & $0.69\scriptstyle{\pm0.00}$ & $\mathbf{0.83\scriptstyle{\pm0.01}}$ & $0.67\scriptstyle{\pm0.02}$ & $0.47\scriptstyle{\pm0.01}$ & 0.62\\
& kNN & $0.45\scriptstyle{\pm0.05}$ & $0.66\scriptstyle{\pm0.04}$& $0.56\scriptstyle{\pm0.13}$ & $0.69\scriptstyle{\pm0.03}$ & $0.63\scriptstyle{\pm0.04}$ & $0.67\scriptstyle{\pm0.06}$ & 0.61\\ 
  & GOL \cite{lee_geometric_2022} & $0.48\scriptstyle{\pm0.08}$ & $0.68\scriptstyle{\pm0.03}$ & $0.59\scriptstyle{\pm0.09}$ & $0.75\scriptstyle{\pm0.03}$ & $0.73\scriptstyle{\pm0.03}$ & $0.62\scriptstyle{\pm0.08}$ & 0.64\\
 & GOL+MDR & $0.62\scriptstyle{\pm0.07}$ & $0.73\scriptstyle{\pm0.02}$ & $0.56\scriptstyle{\pm0.08}$ & $0.75\scriptstyle{\pm0.02}$ & $0.73\scriptstyle{\pm0.02}$ & $\mathbf{0.63\scriptstyle{\pm0.07}}$ & 0.67\\
& GOL+FT+MDR & $\mathbf{0.73\scriptstyle{\pm0.03}}$ & $0.63\scriptstyle{\pm0.06}$ & $\mathbf{0.74\scriptstyle{\pm0.01}}$ & $0.78\scriptstyle{\pm0.01}$ & $\mathbf{0.74\scriptstyle{\pm0.03}}$ & $0.61\scriptstyle{\pm0.02}$ & $\mathbf{0.71}$\\
 \bottomrule
  \end{tabular}
  \label{tab:quantitative_vgg}
\end{table*}

\FloatBarrier

\newpage

\section{Biogeographical Regions}

Biomes influence the distribution of tree species, and at individual level their height, crown area, among other characteristics. The DRIFT dataset covers four major biomes: Mediterranean (Spain, France), Continental (France, Slovenia, Slovakia, Denmark), Alpine (Slovenia, Slovakia), Boreal (Denmark).

\begin{figure}[h!]
  \centering
\includegraphics[width=0.65\textwidth,keepaspectratio]{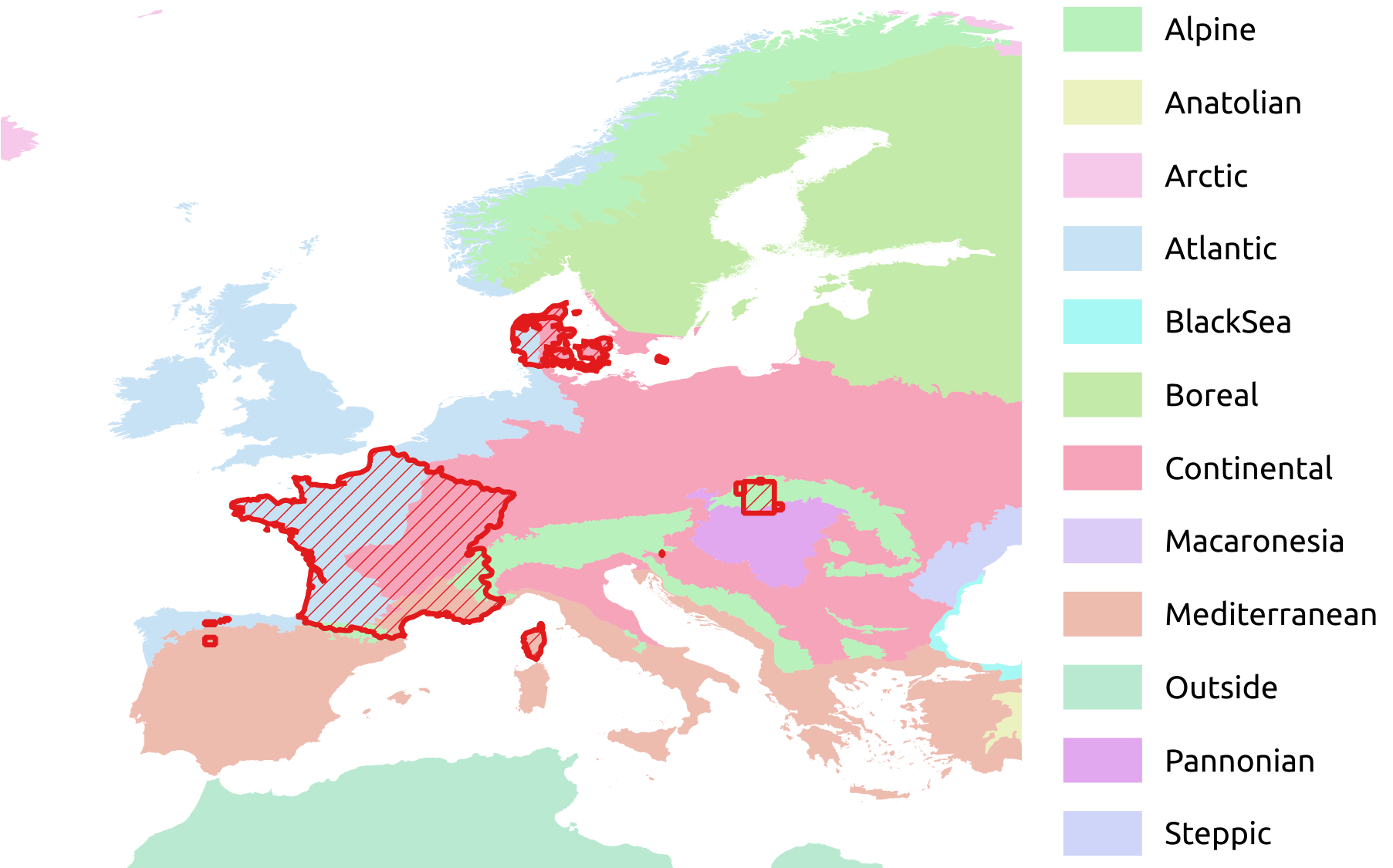}
  \caption{Biogeographical region distribution reflects the diversity of biomes in the DRIFT dataset. The red polygons illustrate the exact locations from which image patches were extracted to build the DRIFT dataset. Image credits: European Environment Agency. \url{https://www.eea.europa.eu/data-and-maps/figures/biogeographical-regions-in-europe-2}
  }
  \label{fig:biom}
\end{figure}

\FloatBarrier
\newpage

\section{Qualitative Results}

\begin{figure}
  \centering
  \includegraphics[width=0.78\textwidth,keepaspectratio]{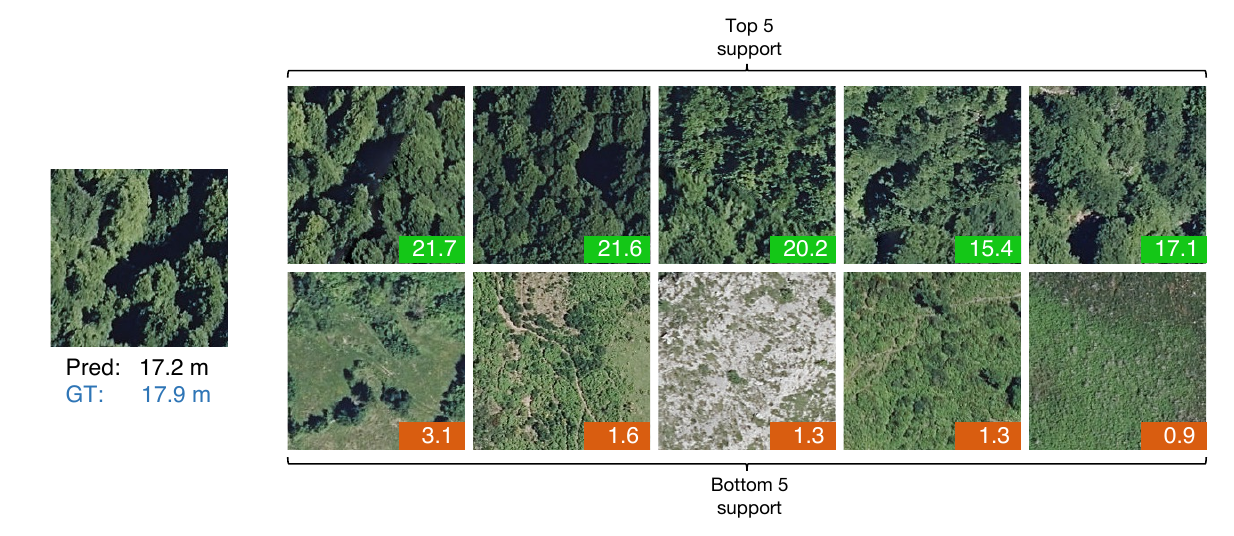}
  \caption{Visual results: five most similar and five least similar support examples for a query, after 5-shot GOL+MDR. Labels are shown in the bottom right corner.
  }
  \label{fig:query}
\end{figure}

\begin{figure}
  \centering
\includegraphics[width=0.78\textwidth,keepaspectratio]{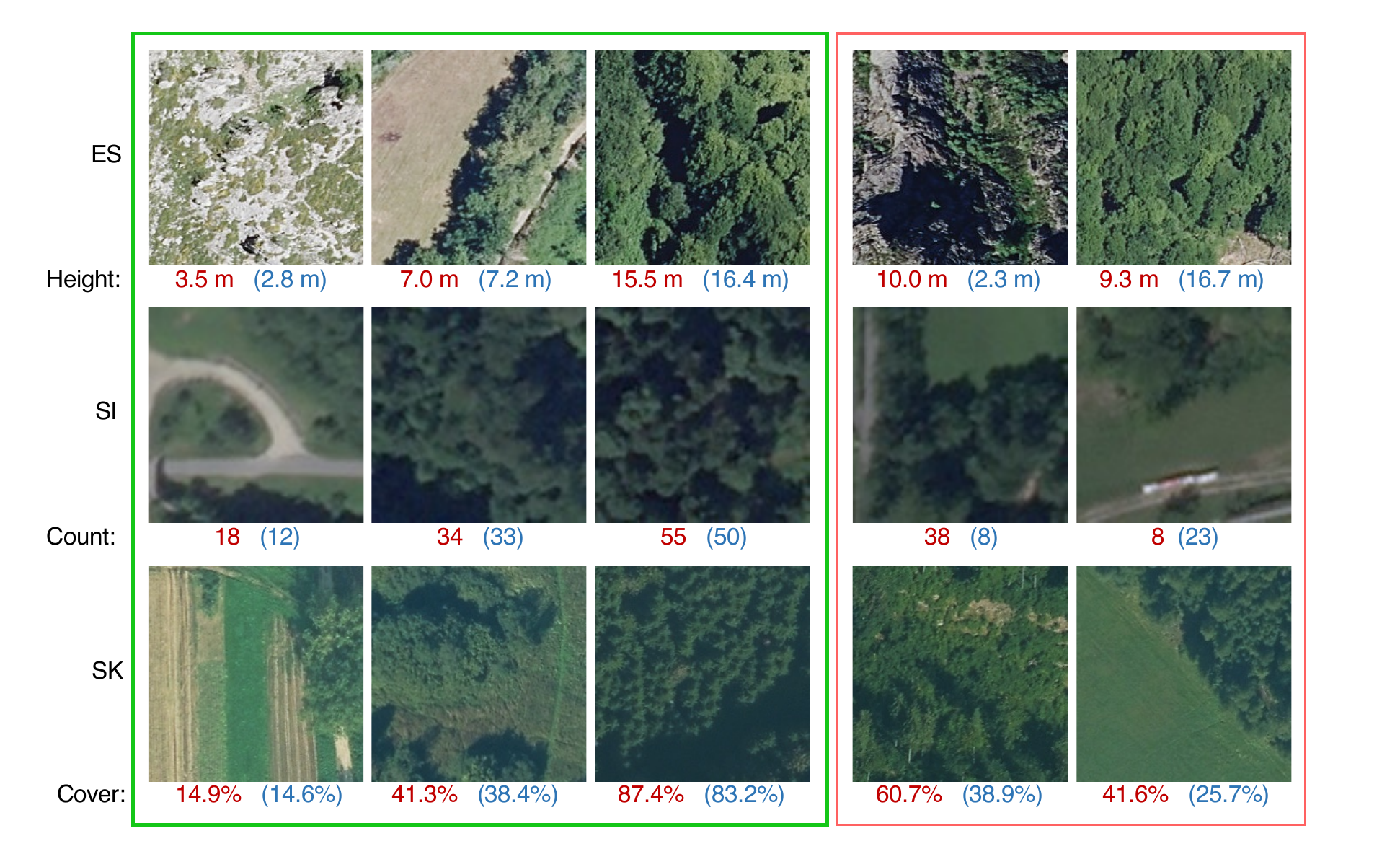}
  \caption{Example predictions by 5-shot GOL+MDR: successful cases on the left (green box) and failure cases on the right (red box). Predictions are shown in red text and ground-truths are shown in blue text in brackets.
  }
  \label{fig:examples}
\end{figure}

\FloatBarrier
\newpage

\section{Effect of Support Set Size}

The performance on the height and count regression tasks increases until 20 shots (20 example per group, 100 in total), then saturates. For the cover task, performance saturates at 10-shots.

\begin{figure}[h]
  \centering
  \includegraphics[width=0.99\textwidth,keepaspectratio]{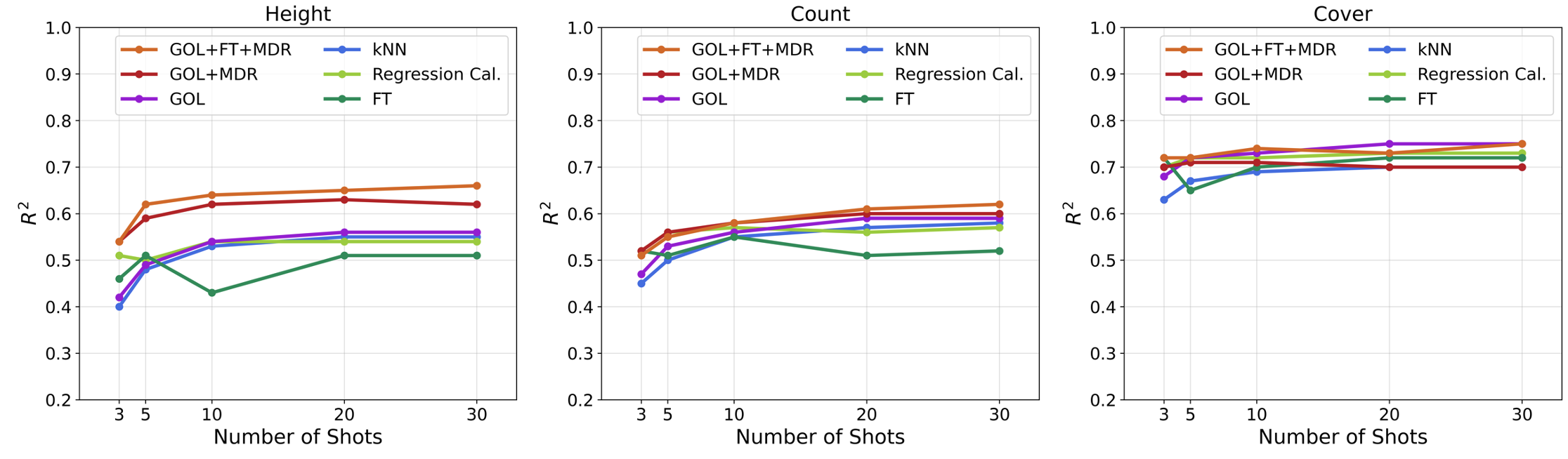}
  \caption{Sensitivity to the number of shots per group. Average $R^2$ score across two source domains and three target domains, with ViT-B/16 backbone.
  }
  \label{fig:shots_nostd}
\end{figure}

\section{MDR Parameters}

As the support set grows larger, the influence of the number of neighbors for kNN assignment diminishes.

\begin{figure}[h]
  \centering
  \includegraphics[height=5cm]{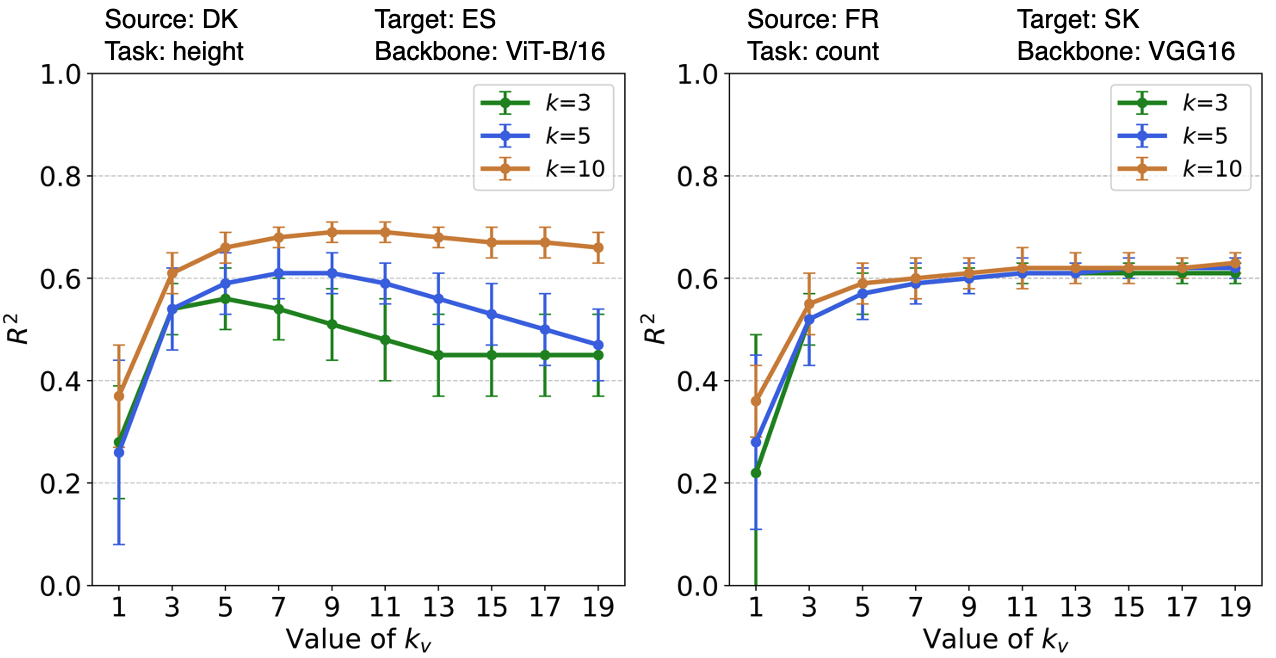}
  \caption{Sensitivity to $k_v$  in the weighted sum for combining relevant support samples. $k=N$ denotes the number of nearest neighbors for constructing the affinity matrix.
  }
  \label{fig:k_values}
\end{figure}

\end{document}